\documentclass{article}

\usepackage{microtype}
\usepackage{graphicx}
\usepackage{subfigure}
\usepackage{booktabs} 
\usepackage[hyphens]{url}

\usepackage{hyperref}
\usepackage{xspace}

\newcommand{\name}{Bandana\xspace}
\newcommand{\company}{Facebook\xspace}

\usepackage[accepted]{sysml2019}

\begin{document}

\twocolumn[
\sysmltitle{\name: Using Non-volatile Memory for Storing Deep Learning Models}




\begin{sysmlauthorlist}
 \sysmlauthor{Assaf Eisenman}{stanford,fb}
 \sysmlauthor{Maxim Naumov}{fb}
 \sysmlauthor{Darryl Gardner}{fb}
 \sysmlauthor{Misha Smelyanskiy}{fb}
 \sysmlauthor{Sergey Pupyrev}{fb}
 \sysmlauthor{Kim Hazelwood}{fb}
 \sysmlauthor{Asaf Cidon}{stanford}
 \sysmlauthor{Sachin Katti}{stanford}
\end{sysmlauthorlist}

 \sysmlaffiliation{stanford}{Stanford University}
 \sysmlaffiliation{fb}{Facebook, Inc}

\sysmlcorrespondingauthor{Assaf Eisenman}{assafe@cs.stanford.edu}

\sysmlkeywords{Non-volatile Memory, Persistent memory, Deep Learning, Deep Learning Storage, Machine Learning, SysML}

\vskip 0.3in

\begin{abstract}
Typical large-scale recommender systems use deep learning models that
are stored on a large amount of DRAM. These models often rely on embeddings,
which consume most of the required memory.
We present \name, a storage system that reduces the DRAM footprint of embeddings, by using Non-volatile Memory (NVM) as the primary storage medium, with a small amount of DRAM as cache.
The main challenge in storing embeddings on NVM is its limited read bandwidth compared to DRAM.
\name uses two primary techniques to address this limitation:
first, it stores embedding vectors that are likely to be read together in the same physical location, using hypergraph partitioning,
and second, it decides the number of embedding vectors to cache in DRAM by simulating dozens of small caches. 
These techniques allow \name to increase the effective read bandwidth of NVM by 2-3$\times$ and thereby significantly reduce the total cost of ownership.
\end{abstract}

]



\printAffiliationsAndNotice{}  

\section{Introduction}

An increasing number of web-scale applications are relying on deep learning models, including
online search~\cite{rankbrain}, online ads~\cite{deeplearning-ads}
and content recommendation systems~\cite{content-recommendation}.
Typically, the precision of deep learning algorithms increases as a
function of the model size and the number of features.
Therefore, application providers are dedicating ever more compute and storage resources
to training, storing and accessing deep learning models.

For example, at \company,
thousands of servers are dedicated to storing deep learning models to recommend relevant posts or content to users.
The deep learning features are often represented by vectors called embeddings, which encode the meaning
of each feature, such that similar vectors are closer in the embedding Euclidean space.
To compute the most relevant posts to serve each user, \company uses two types of embeddings:
user and post embeddings. Post embeddings represent the features of the post themselves (e.g.,
the main words), while the user embeddings represent features unique to each user, which
represent their topics of interest and past activity.
At \company, both types of embeddings are fully stored in DRAM, in order to enable real-time access
for computing the most relevant content for each user.

However, DRAM is a relatively expensive storage medium, and in fact has gotten even more expensive recently,
due to shortages in global supply~\cite{dram-exchange,marketwatch-dram}. In this work, our goal is to
minimize the amount of DRAM used to store embeddings, and therefore the total cost of ownership (TCO).
Since post embeddings need to go through more ranking and processing, they have a much longer pipeline
than user embeddings. Therefore, user embeddings can be read early in the process
from a slower but cheaper storage medium than DRAM.

Non-volatile Memory (NVM), also termed Non-volatile Main Memory or persistent memory, offers an attractive storage medium
for user embeddings, since it costs about an order of magnitude less per bit
than DRAM, and its latency and throughput can satisfy the requirements of computing user embeddings.
However, even though NVM provides sufficient latency to meet the system's requirements,
its bandwidth is significantly lower than DRAM. Exarcerbating the problem,
NVM devices offer maximum bandwidth only if
the size of reads is 4~KB or more, while user embedding vectors are only 64-128~B.
Therefore, na\"{\i}vely substituting DRAM for NVM
results in underutilized bandwidth, and causes both its latency to increase and the application's throughput to drop significantly.

We present \name, an NVM-based storage system for embeddings of recommender systems, which optimizes the read bandwidth of
NVM for accessing deep learning embeddings. \name uses two primary mechanisms to optimize bandwidth:
storing embedding vectors that can be prefetched together to DRAM, and deciding which vectors to cache in DRAM to maximize
NVM bandwidth.

\paragraph{Prefetching embedding vectors.} Our NVM device benchmarks show that to optimize bandwidth,
NVM needs to be read at the granularity of a 4~KB block or more.
Therefore, \name stores multiple embedding vectors that are likely
to be accessed together in the same 4~KB NVM block, and when one of the objects needs to be read,
it has the option of prefetching the entire block to DRAM. We evaluate two techniques for partitioning the vectors
into blocks: Social Hash Partitioner (SHP)~\cite{SHP}, a supervised hypergraph partitioning algorithm that 
maximizes the number of vectors in each block that were accessed in the same query,
and K-means \cite{KMEANS,KMEANSpp}, an unsupervised algorithm that is run recursively in two stages.
We found that SHP doubled the effective bandwidth compared to K-means for some workloads.

\paragraph{Caching vectors in DRAM.} Even after storing related
vectors together, some blocks contain vectors that will not be read and can be discarded.
Therefore, \name decides which vectors to keep in DRAM, by using a Least Recently Used (LRU) queue,
and only inserting objects from pre-fetched blocks to the queue that have been accessed $t$ times in the past.
We find that the performance varies widely across different embedding tables
based on the value of $t$ and the cache size. Therefore, inspired by recent research
in key-value caches~\cite{mini-caches}, \name runs dozens of
``miniature caches'' that simulate the hit rate curve of different values of $t$ for each embedding table, with a very low overhead.
Based on the results of the simulations, \name picks  
the optimal threshold  
for each embedding table.

We demonstrate that \name significantly improved the effective bandwidth of NVM, enabling it to be used as a primary storage medium for embeddings.
To summarize our contributions:
\begin{enumerate}
\itemsep-0.1em 
\item To our knowledge, \name is the first system that leverages NVM to store deep learning models. It is also one of the first
published systems to use NVM in large-scale production workloads.
\item Using the past access patterns of embedding vectors, \name applies hypergraph partitioning
to determine which vectors are likely to be accessed together in the future.
\item Applies recent techniques from key-value caching to
run lightweight simulations of dozens of miniature caches to determine how aggressively
to cache prefetched vectors for different cache sizes.
\end{enumerate}

\section{Background}

In this section we provide background on two topics:
deep learning embedding vectors and how they are used at \company for recommending posts, as well as NVM.

\subsection{Embedding Vectors}
The goal of \company's post recommendation system is to recommend relevant content to users.
A straightforward way to train a ranking system for this purpose, would be to encode
the post and user features and use them for predicting the likelyhood of a click. 
For example, we can represent users based on the pages they liked. In this case,
each page would correspond to a unique ID, and IDs would be mapped to an element index in a binary vector. 
For instance, if a user liked pages with the IDs 2 and 4, out of a total number of 5 pages,
the user feature vector would be (0,0,1,0,1). 

On the other hand, we could represent posts based on the words used to describe them. In this scenario,
each word would correspond to a unique ID, and once again IDs would be mapped to an element index in a binary vector.
For example, if the post's text is ``red car'' out of word dictionary ``bicycle, mototrcycle, car, blue, red, green'',
the post feature vector would be (0, 0, 1, 0, 1, 0).

Since the total number of words in different languages is on the order of millions, while the number of pages is on the order of billions, 
such binary vectors would be very large and sparse (i.e., will contain mostly zeros). 
Moreover, such a na\"{\i}ve representation would not apply any contextual information of one page to another.
In other words, very similar pages would still be encoded as separate IDs.

However, instead of representing each word or page by a single binary digit, if we represent them by a short vector,
we could represent the similarity between words or pages. The mapping of items into a vector
space is called an embedding.
Embeddings are learned and created in such a way that sparse IDs with similar meaning (in terms of semantics, engagement, etc.)
will also be located closer in terms of distance. The distance is often measured in Euclidian space.
Recommending posts is a specific use case of recommender systems~\cite{content-recommendation,wide-deep,ad-click-prediction}, in which
the goal is to recommend the most relevant content based on past user behavior. 

\begin{figure}[t]
 \begin{center}
  \includegraphics[width=0.6\columnwidth]{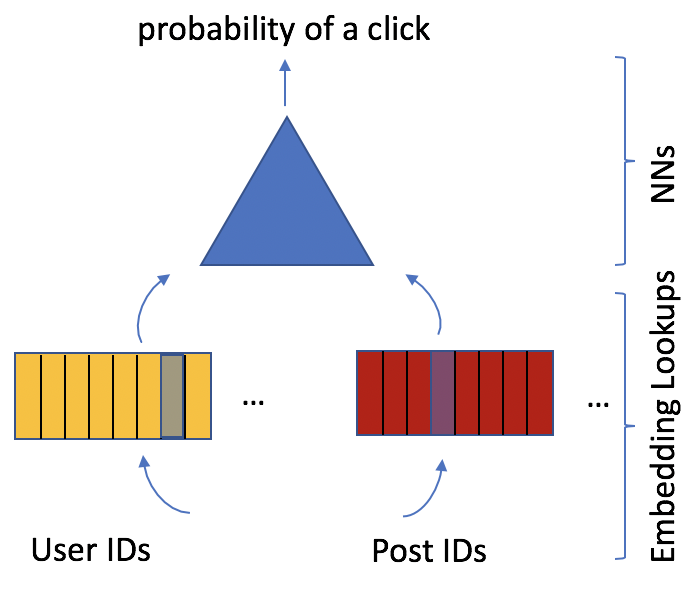}
  \vspace{-1\baselineskip}
  \caption{A deep learning recommendation model.}
  \label{fig:hypothetical_model}
 \end{center}
 \vskip -0.3in
\end{figure}

At \company, embedding vectors are stored in dedicated tables,
where each column represents an embedding vector,
and its column ID corresponds to its ID.
\company maintains two types of embeddings: user and post embeddings.
Each user embedding table can typically represents some kind of user behavior,
such as clicks, likes, and page views, and each embedding
vector is a specific action taken by the user.
The post tables can represent the actual content of the post, such as the specific
phrases appearing in the post.
The recommendation model receives input IDs, extracts the corresponding 
embeddings, and processes them with deep neural networks (NNs), as shown on Figure~\ref{fig:hypothetical_model}. 

The typical vector dimension
in our models is between 32 to 64, where each element occupies 1-4 bytes.
Embedding tables can contain tens of millions of embedding vectors,
requiring on the order of GBs per table.
Due to their latency requirements, these tables are usually stored in DRAM.

The embedding vectors are computed during training, where for each data sample 
(e.g. user and post pair) only the columns accessed by the corresponding IDs are modified. 
Therefore, as the training proceeds through the dataset most (if not all) columns are updated, 
many of which are updated multiple times. The embeddings are then used without any adjustments or
modifications during inference. The vectors may be retrained every few hours.

\subsection{NVM}
\label{sec:background-nvm}

Non-volatile Memory (NVM), is a new memory technology that provides much lower latency and higher throughput than flash, but
lower performance than DRAM, for a much lower cost than DRAM.
NVM can be used in two form factors: the DIMM form factor, which is byte-addressable,
or alternatively as a block device.
The DIMM form factor is currently not supported by Intel processors~\cite{dimm-news}, and is much more expensive
than using NVM as a block device. Therefore, for our use case, we focus on the block device form factor.

To understand how to use NVM, we explored its performance characteristics. For this purpose, we ran a widely
used I/O workload generator, Fio 2.19~\cite{fio}, on an NVM device.
We ran the Fio workloads with 
4 concurrent jobs using the
Libaio I/O engine with different queue depths. The queue depth represents the number of outstanding I/O requests to the device,
which is a proxy for how many threads we run in parallel. We measured the latency and bandwidth of an
NVM device with a capacity of 375~GB. 

\begin{figure}[t!]
 \begin{center}
  \subfigure[]{\label{fig:nvm_avg_latency}\includegraphics[width=.49\columnwidth]{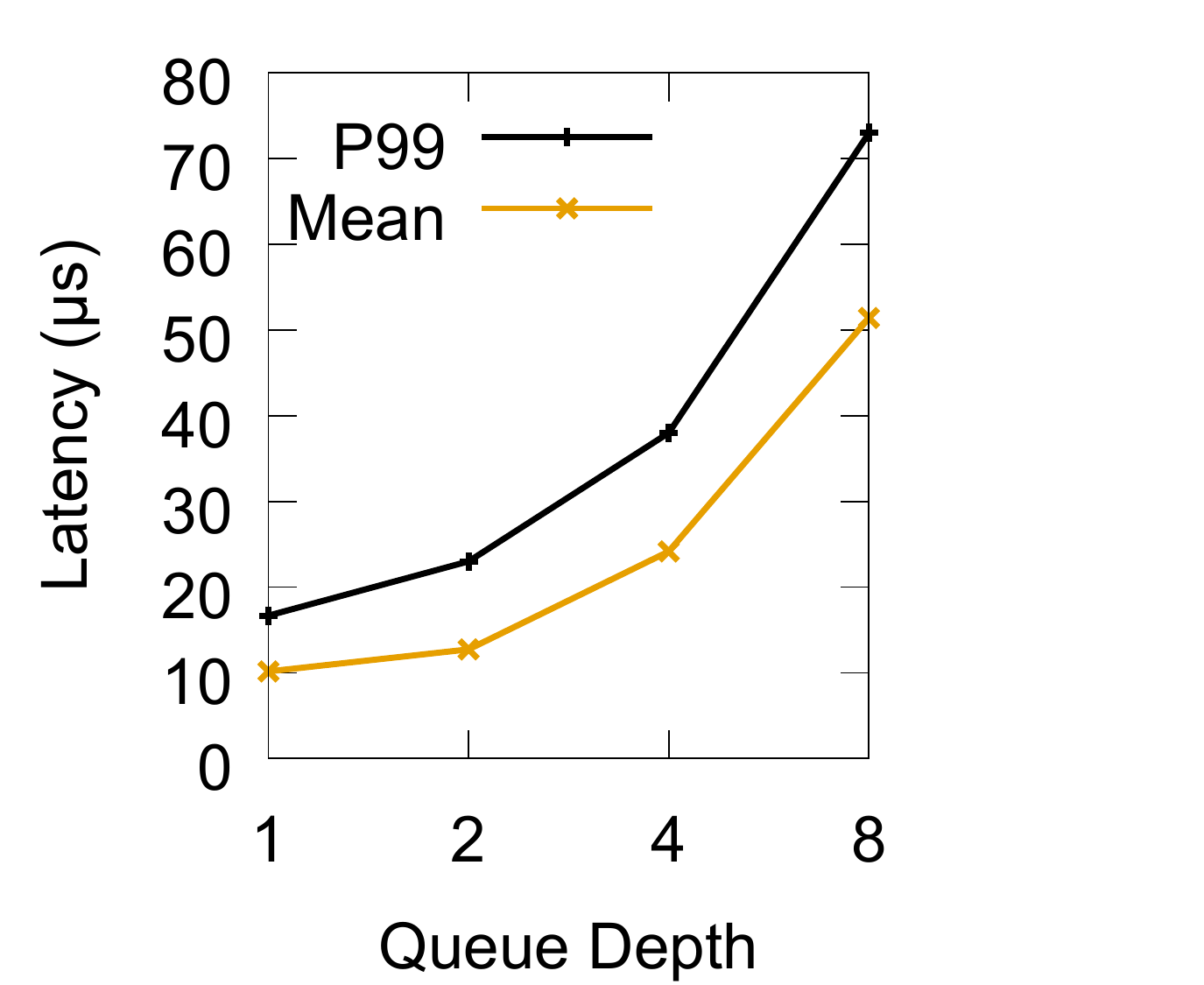}}
  \subfigure[]{\label{fig:nvm_bw}\includegraphics[width=.49\columnwidth]{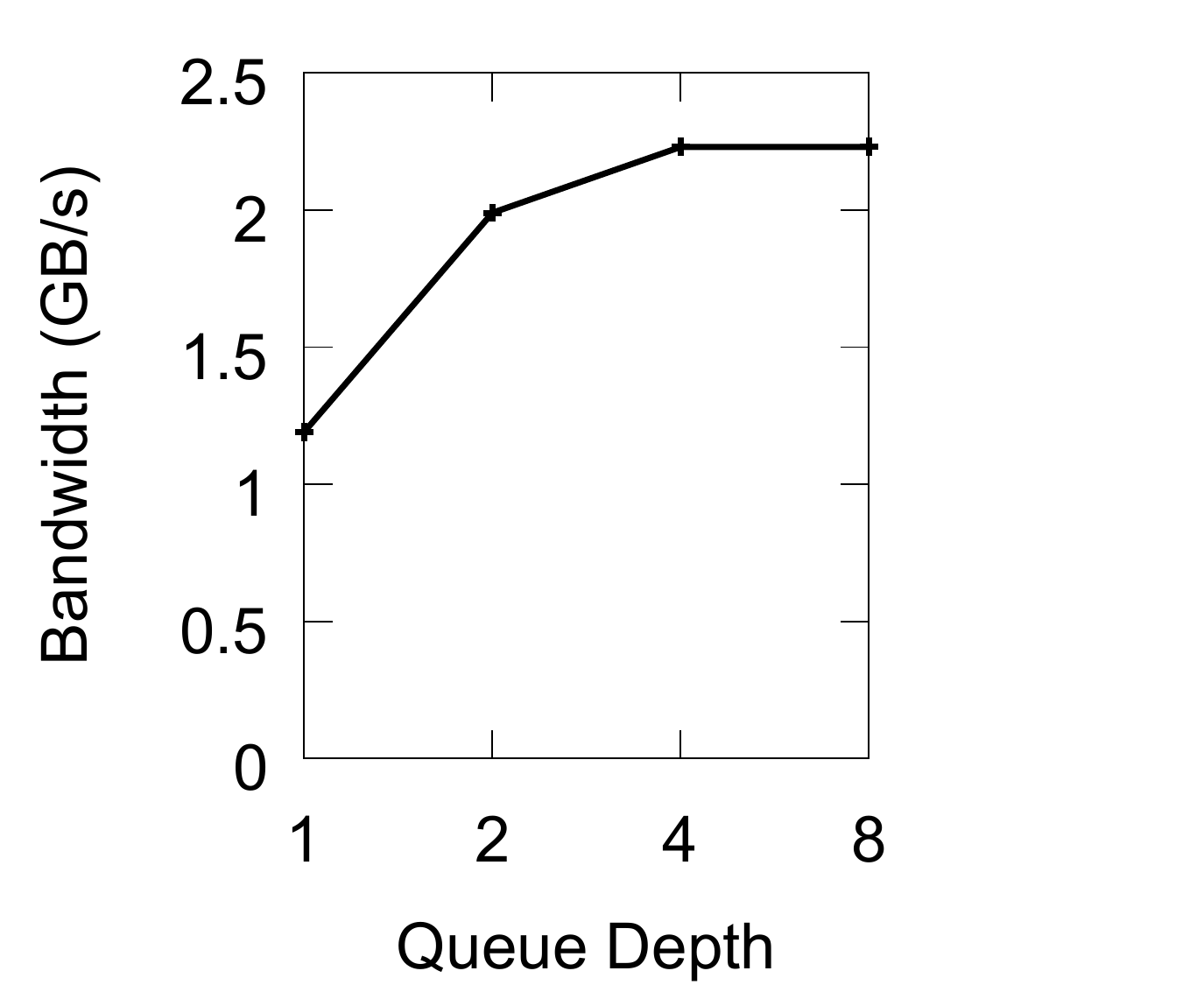}}
  \vskip -0.1in
  \caption{The latency and bandwidth for a 4KB random-read workload with variable queue depths.}
  \label{fig:fio}
 \end{center}
 \vskip -0.3in
\end{figure}

Figure~\ref{fig:fio} presents the average latency, P99 latency, and bandwidth for a read-only workload with random accesses of 4~KB.
The results show that there is a trade-off between latency and bandwidth: a higher queue depth
provides higher read bandwidth, at the expense of higher latency.
Note that even at a high queue depth, NVM's read bandwidth (2.3~GB/s) is still $>30\times$ lower than DRAM (e.g., 75~GB/s).

Note that unlike DRAM, NVM's endurance deteriorates as a function of the number of writes over its lifetime.
Typical NVM devices can be re-written 30 times a day, or they will start
exhibiting errors. Fortunately, the rate of updating the vectors at \company is often between 10-20 times a day, which is below the rate
that would affect the endurance of the NVM devices.

\section{Workload Characterization}
This section presents a workload characterization of the user embeddings at \company. 
We analyze a production workload containing 1 billion embedding vector lookups, representing traffic 
of over one hour for a single model.
Currently, the number of models per server and the size of each 
model are bounded by the DRAM capacity of the server.

Each user embedding table typically represents a different class of user behavior.
For example, a table might represent pages liked by the user, where each embedding vector represents a page. 
Hence, a request in \name usually incorporates multiple tables and contains multiple vector lookups
inside each table. Because different posts are ranked for a single user, the post embeddings are read 
much more frequently, and post lookups constitute about 95\% of the total embedding reads. On the other 
hand, user embeddings contain more features, and consume about 75\% of the total DRAM capacity. 

In the model we analyze, embedding vector are 128 bytes containing 64 elements of type \texttt{fp16}.
Table~\ref{characterization-table} describes the characteristics of some representative user embedding tables in the model.
Each embedding table is comprised of 10-20 million vectors (between 1.2~GB to 2.4~GB). The average number of vectors included in a 
single request varies across the tables, with 17.68 vector lookups (on average) in embedding table 8, and up to 92.8
vector lookups (on average) in embedding table 2. 
The table also presents the vector lookup distribution across the user embedding tables. 
The largest part of vector lookups is consumed by embedding table 2, which serves 25\% of the 
user embedding lookups.
Compulsary misses describe how many of these lookups were unique (i.e., how many lookups correspond to 
vectors that were not read before in the trace). The lower the percentage of compulsory misses, the more likely the table can be effectively cached.

\begin{table}[t]
\caption{Characterization of the user embedding tables.}
\label{characterization-table}
\vskip -0.1in
\begin{center}
\begin{small}
\begin{sc}
\setlength{\tabcolsep}{0.3em}
\begin{tabular}{lcccc}
\toprule
Table  & Vectors & Avg request  & \% of total & Compulsary \\
           &                &  Size           & lookups & misses \\
\midrule
1  & 10M & 34.83 &  9.44\% & 4.16\% \\
2  & 10M & 92.75 &  25.14\% & 2.19\% \\
3  & 20M & 26.67 &  7.23\% & 24.29\% \\
4  & 20M & 25.14 &  6.82\% & 19.46\% \\
5    & 10M & 30.22 & 8.19\% & 22.68\% \\
6    & 10M & 53.50 &  14.5\% & 26.94\% \\
7    & 10M & 54.35 & 14.73\% & 11.36\% \\
8  & 20M & 17.68 &  4.79\% & 60.83\% \\
\bottomrule
\end{tabular}
\end{sc}
\end{small}
\end{center}
\vskip -0.3in
\end{table}

\begin{figure*}[t]
  \begin{center}
   \subfigure[Table 1]{\includegraphics[width=.245\textwidth]{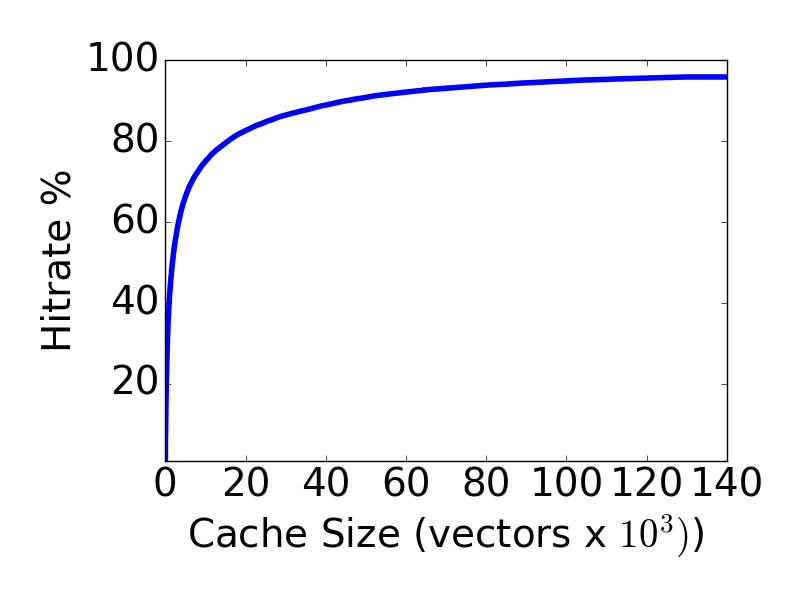}}
   \subfigure[Table 2]{\includegraphics[width=.245\textwidth]{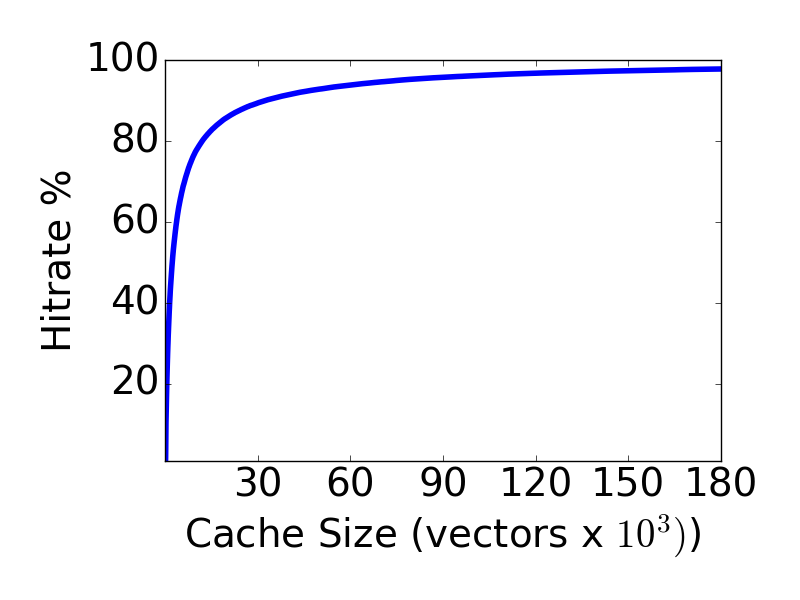}}
   \subfigure[Table 6]{\includegraphics[width=.245\textwidth]{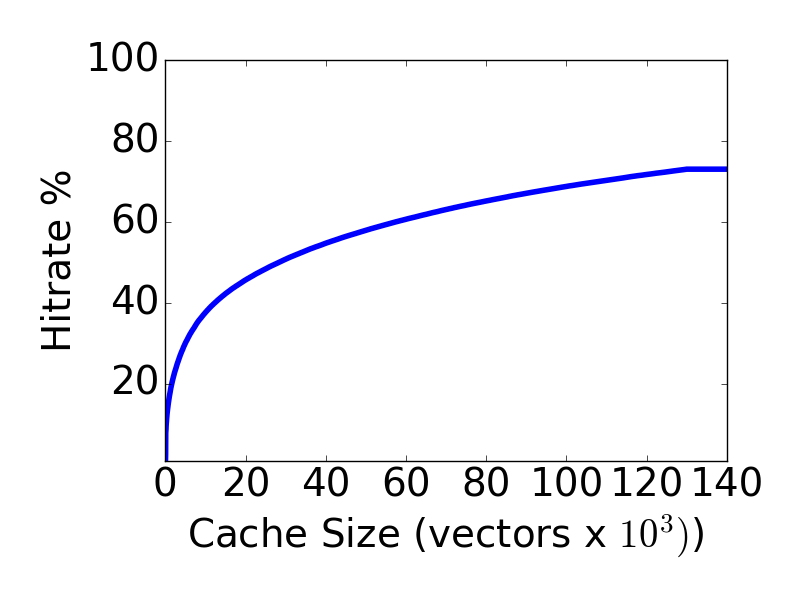}}
   \subfigure[Table 7]{\includegraphics[width=.245\textwidth]{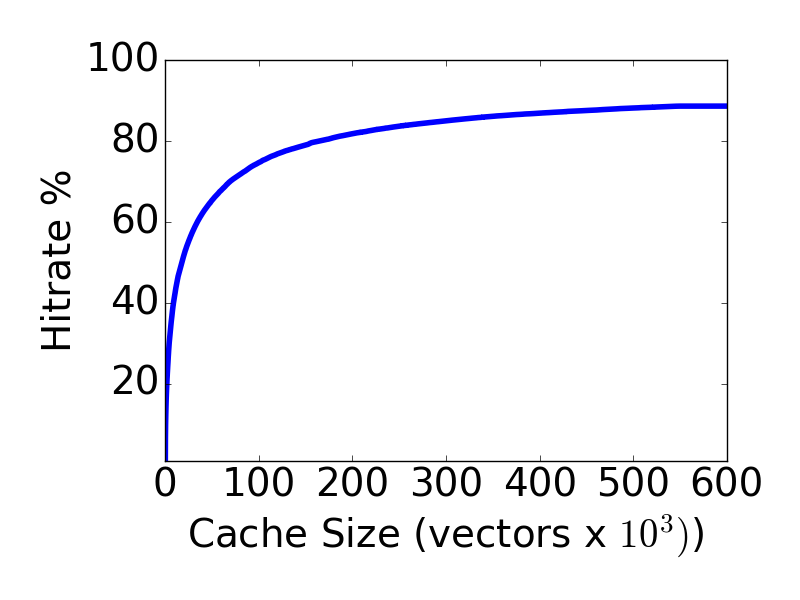}}
   \vskip -0.1in
   \caption{Hit rate curves of the user embedding tables with the top number of lookups.}
   \label{fig:stack_distances}
  \end{center}
  \vskip -0.3in
\end{figure*}

\begin{figure*}[t]
  \begin{center}
   \subfigure[Table 1]{\includegraphics[width=.245\textwidth]{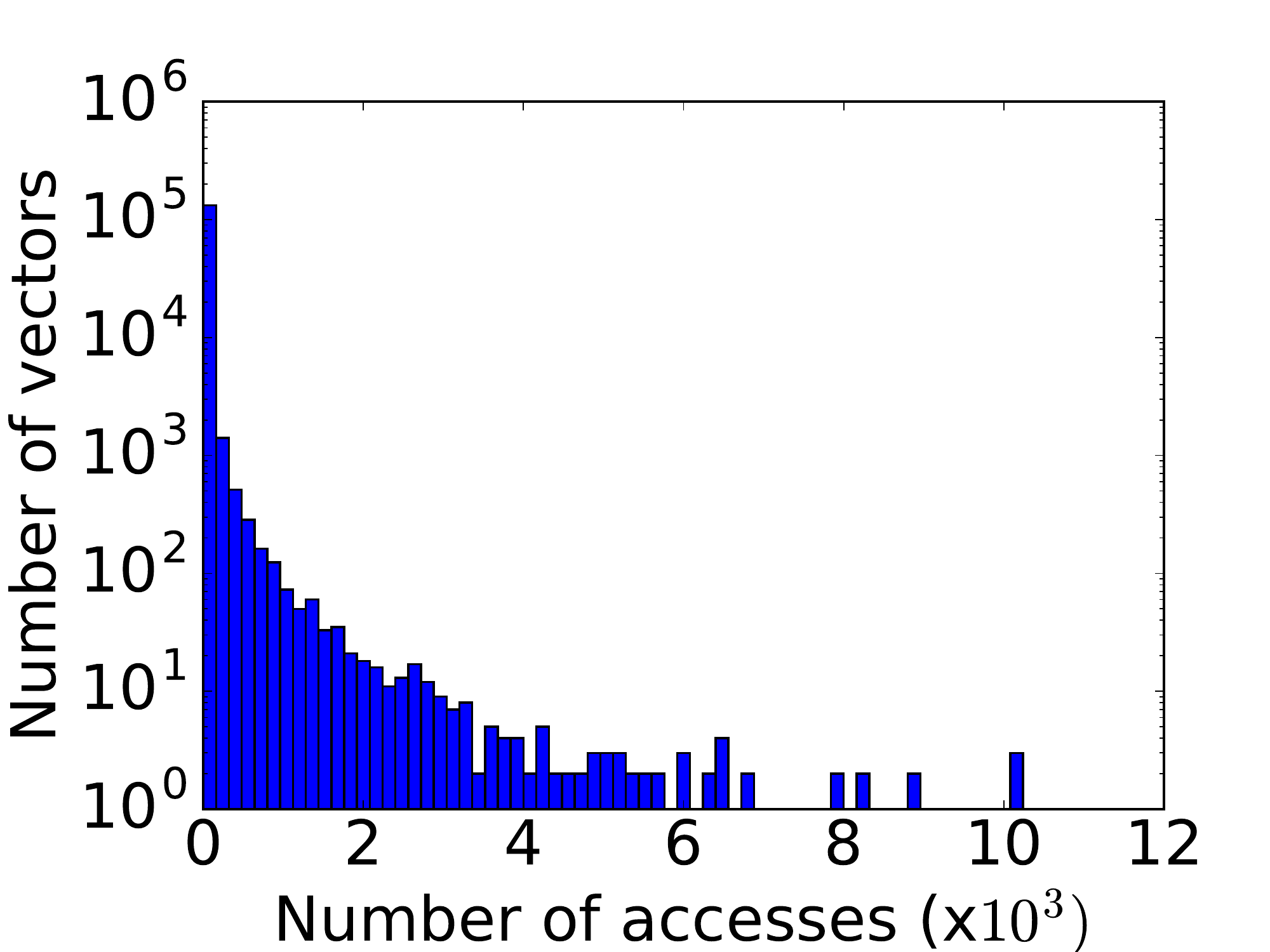}}
   \subfigure[Table 2]{\includegraphics[width=.245\textwidth]{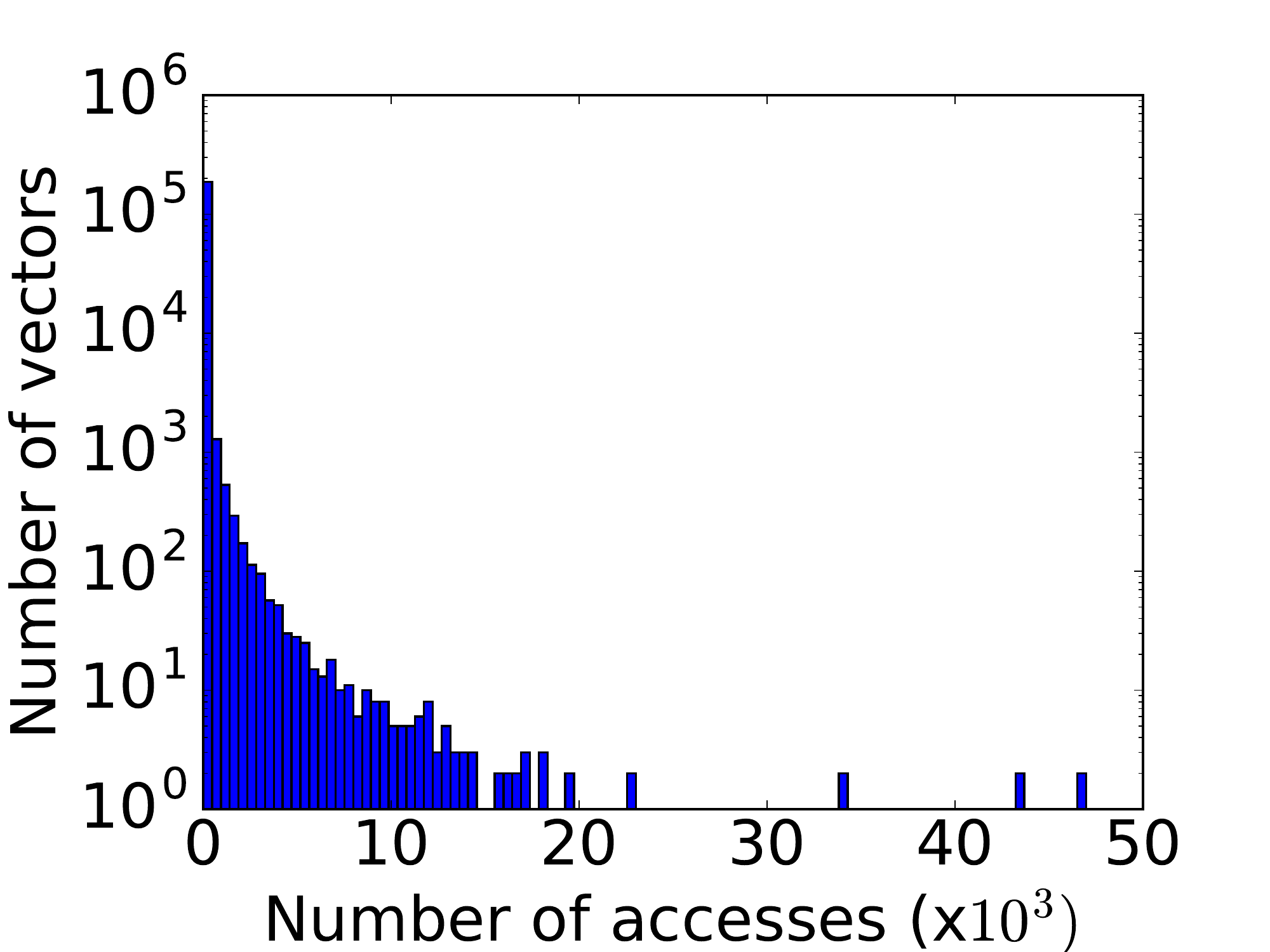}}
   \subfigure[Table 6]{\includegraphics[width=.245\textwidth]{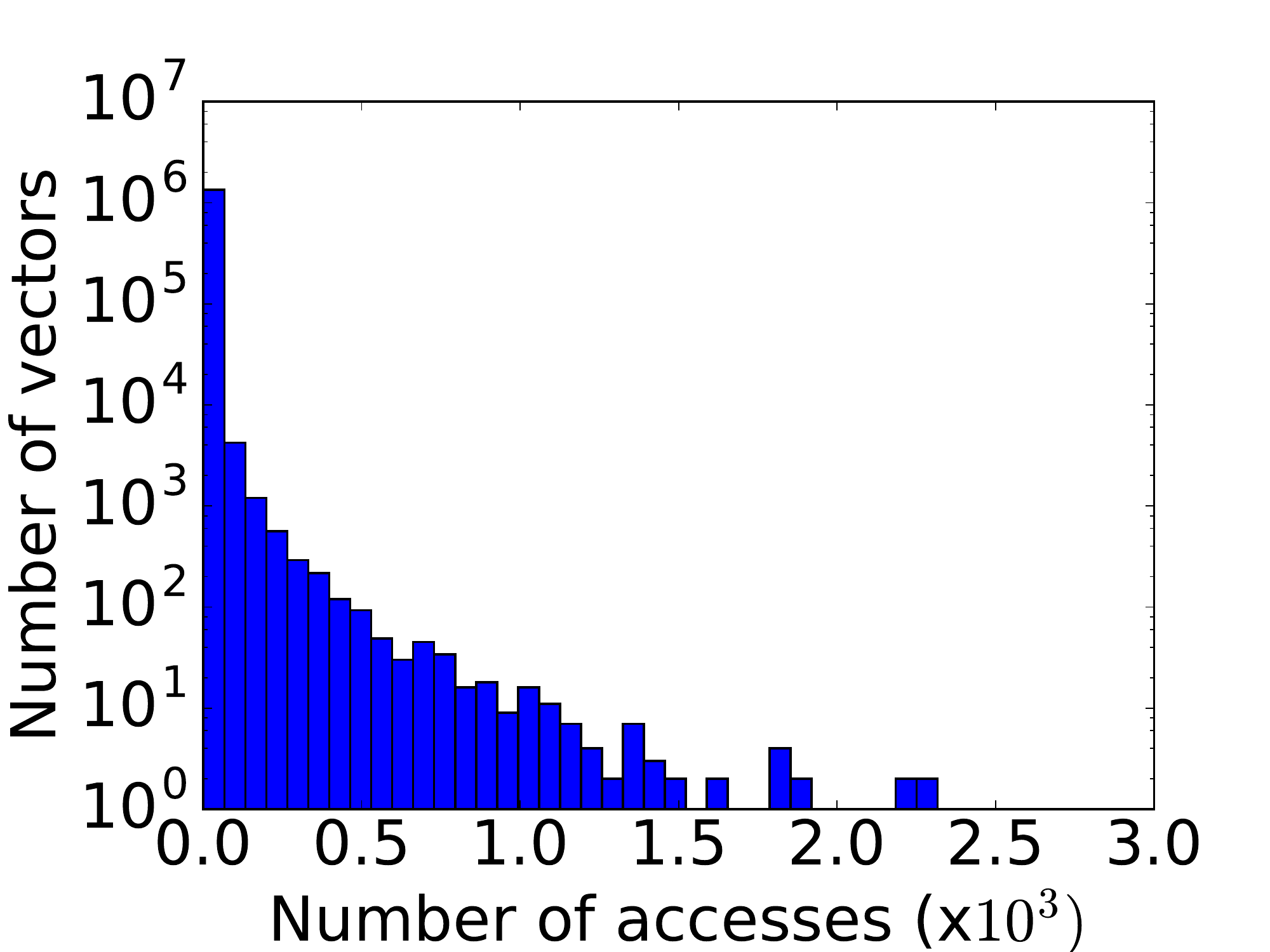}}
   \subfigure[Table 7]{\includegraphics[width=.245\textwidth]{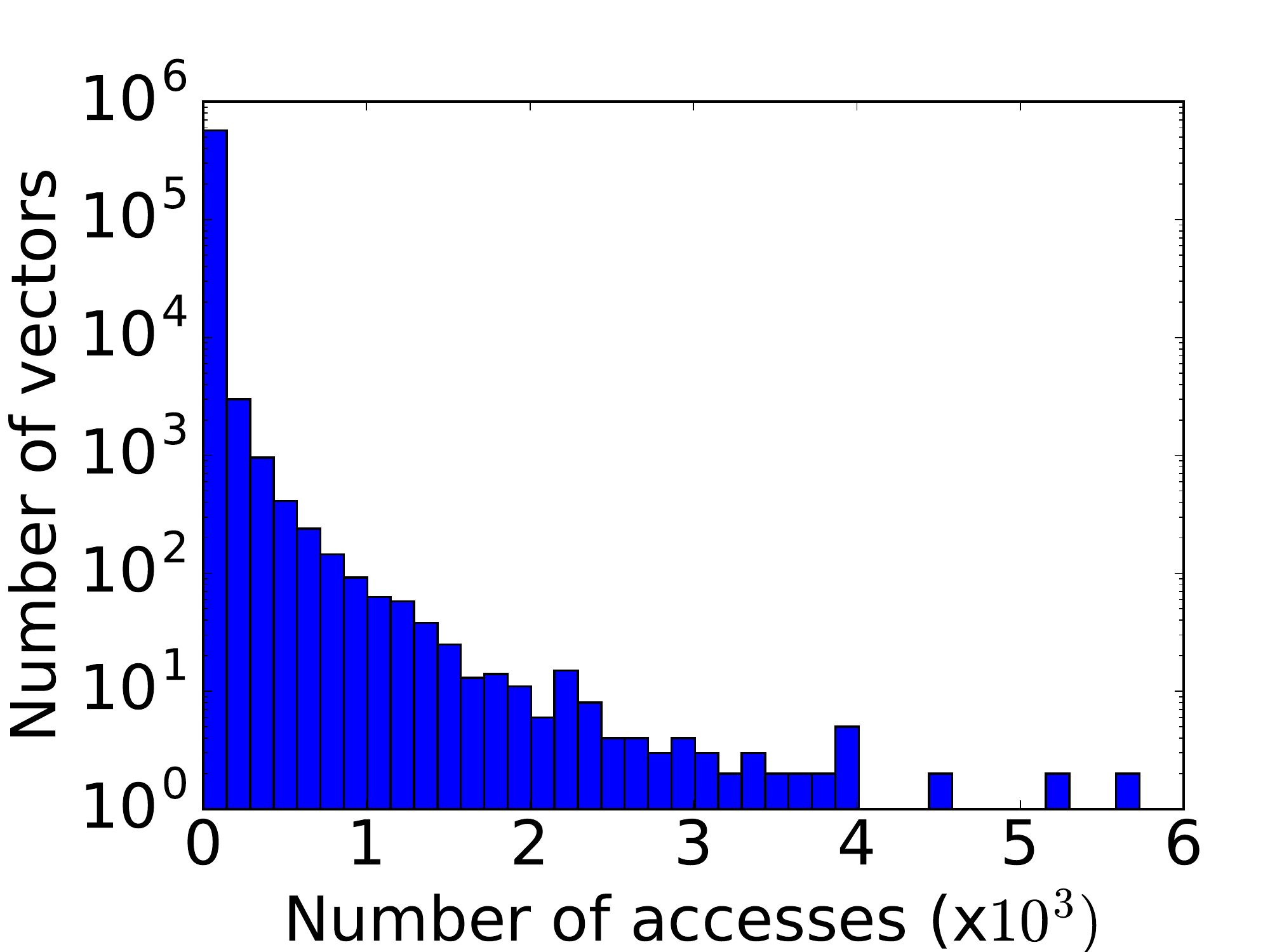} }
   \vskip -0.1in
   \caption{Access histograms of the user embedding tables with the top number of lookups.}
   \label{fig:accesses}
  \end{center}
  \vskip -0.2in
\end{figure*}

To gain more insight on the reuse of the user embedding vectors, we calculate the stack 
distances~\cite{stack_distance} of each embedding table. To compute them, we assume each table is cached in an infinite LRU queue,
where the stack distance of a vector is its rank 
in an LRU queue at the time it is requested, counted from the top of the eviction queue. For example, if the requested vector 
is at the second object in the eviction queue when it is requested, its stack distance is equal to 2. 
This allows us to compute the hit rate curve as a function of the memory allocated to embedding table.  
Figure~\ref{fig:stack_distances} depicts the hit rate curves of the four embedding tables with the top number of lookups,
in a trace of one billion requests.
Figure~\ref{fig:accesses} shows the access histogram of these tables, where each bar depicts how many vectors
(X axis) were read a certain number of times (Y axis).  
The histograms show that there is a very high variance in the access patterns of the tables. For example,
table 2 contains vectors that are read 100,000s of times, while for table 7 there are no vectors that
are read more than 1,000 times.

\section{Design}
This section presents the design choices and trade-offs when designing an NVM-based storage system for embedding tables.

\subsection{Baseline}

A simple approach for using NVM to store recommender system embedding tables is to cache
a single vector that is read by the application in DRAM, and evict a single old vector at a time.
We refer to this policy throughout the paper as the \emph{baseline policy}.

\begin{figure}[t!]
 \begin{center}
  \subfigure[]{\label{fig:nvm_avg_latency}\includegraphics[width=.49\columnwidth]{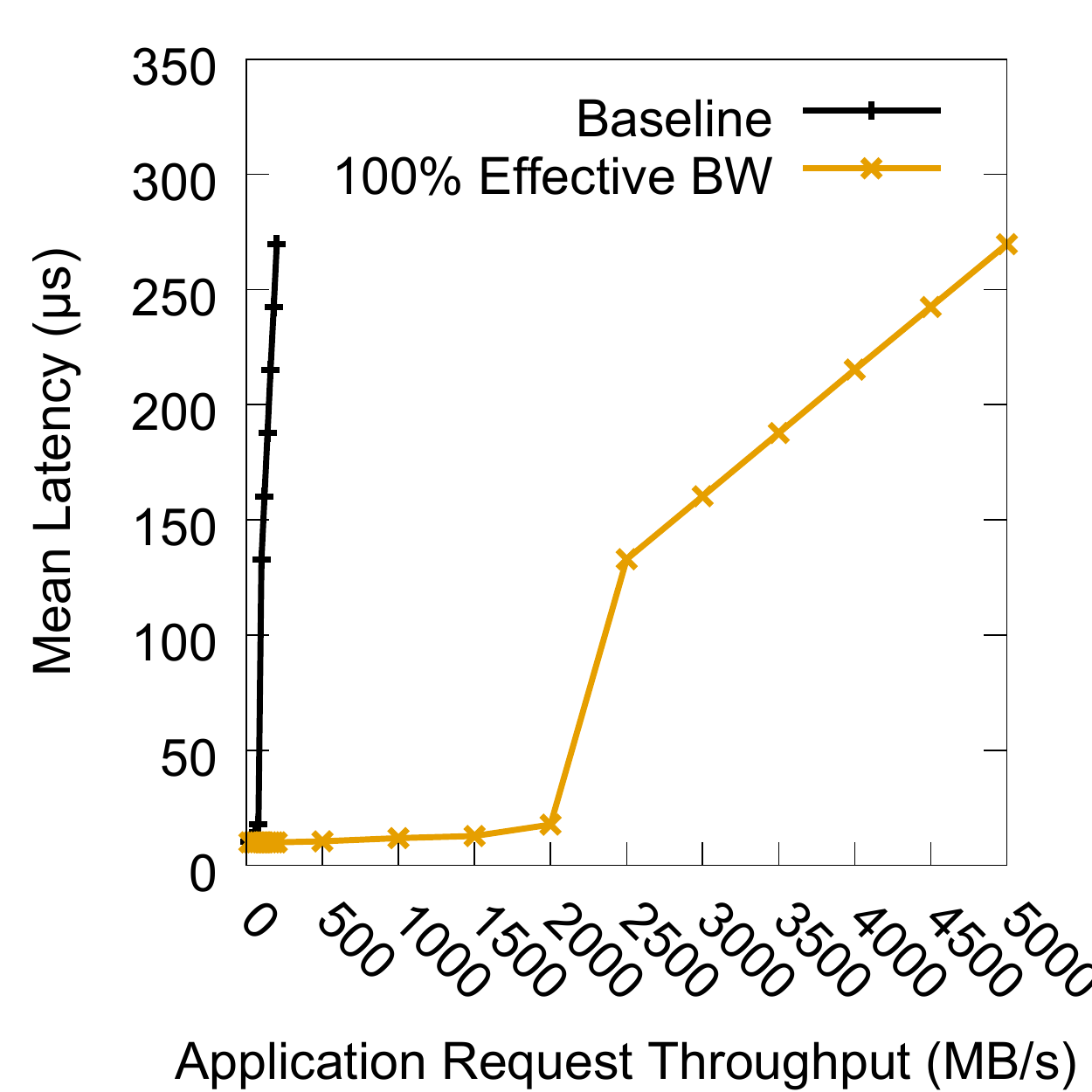}}
  \subfigure[]{\label{fig:nvm_bw}\includegraphics[width=.49\columnwidth]{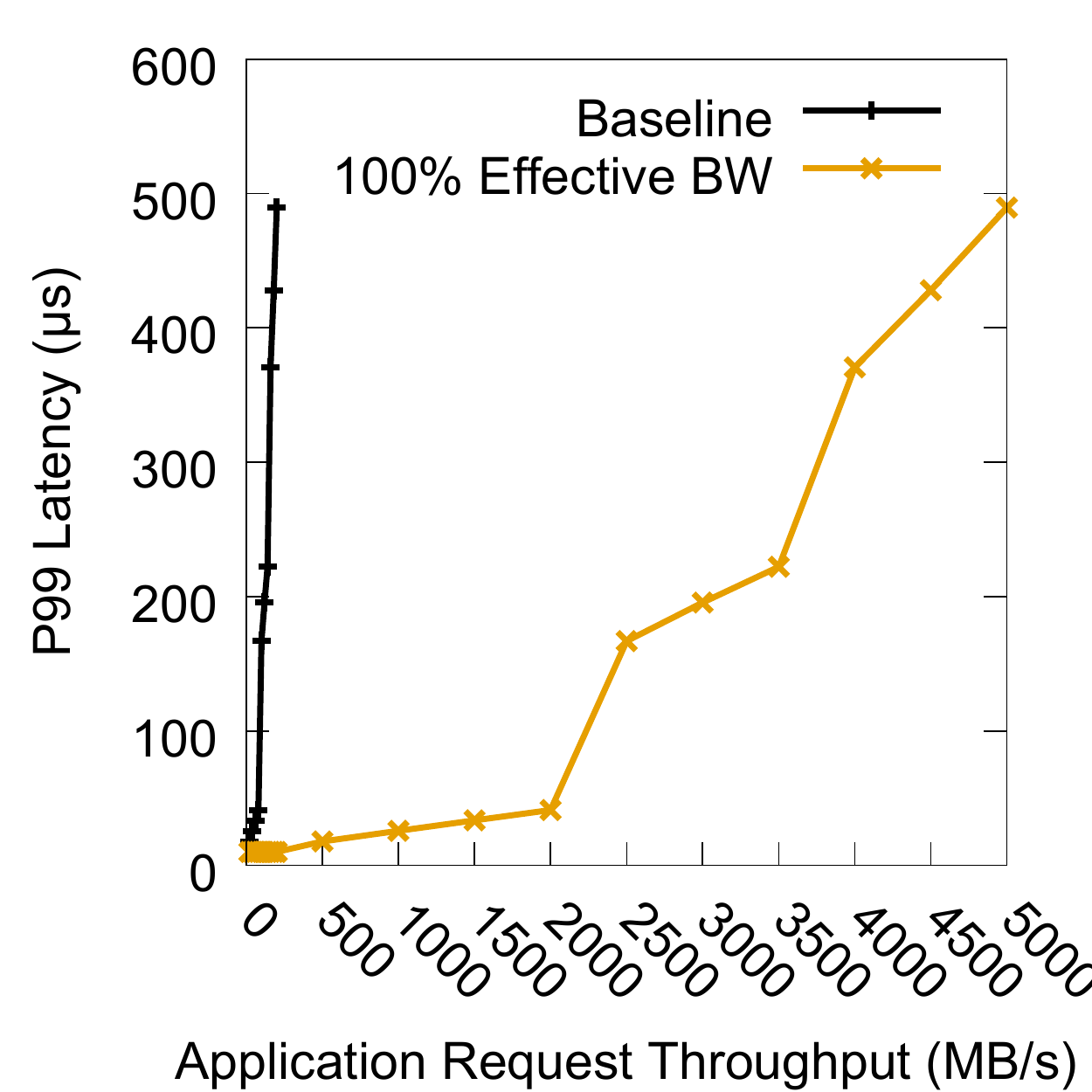}}
  \vskip -0.1in
  \caption{Mean and P99 latencies as a function of the throughput of a 375~GB NVM device.
  The baseline policy represents the scenario where the application issues a 128~B NVM read for each embedding
  vector. The 100\% effective bandwidth line represents the performance of the NVM device with random 4~KB reads.}
  \label{fig:latency_emulation}
 \end{center}
 \vskip -0.2in
\end{figure}

Figure~\ref{fig:latency_emulation} presents the latency as a function of the throughput of the NVM device for the baseline policy, as well
as for a synthetic workload that issues random 4~KB reads from the NVM.
The reason that latency under the baseline policy as a function of throughput is much higher than that of a device
where we issue 4~KB reads, is due to the fact that NVM reads in the block device form factor are in the 
granularity of 4~KB blocks, while the size of an embedding vector is only 128~B. Therefore, the baseline policy is 
not utilizing more than 96\% of the read bandwidth of the NVM device. 
We use the term \emph{effective bandwidth} to denote the percentage of NVM read bandwidth that is read by the application.
In the case of the baseline policy, the effective bandwidth is only 4\% of the total bandwidth of the NVM, and the rest is discarded.
Therefore, under a high load, when the effective bandwidth is so low, the latency of NVM spikes (and throughput drops).

Instead of reading a single vector to DRAM when it is accessed, an alternative
approach would be to read all 32 vectors stored in its physical 4~KB block.
However, when we use a limited cache size, the policy of caching all 32 vectors that belong to a block,
performs even worse than fetching one vector at a time,
and reduces the effective bandwidth by more than 90\% compared to the baseline policy.
This is due to the fact that the vectors stored in the same physical block as the requested vector are not read before they are evicted, since
they have no relationship with the vector that had just been read.
Therefore, fetching them together with the requested vector offers no benefit (in \S\ref{sec:caching}
we further analyze the performance of caching the prefetched vectors without ordering them).

In summary, the limited effective bandwidth is the main bottleneck
to adopting NVM as an alternative for DRAM for storing embeddings.
Therefore, the main objective of \name is to maximize the  
 \emph{effective bandwidth increase} 
over the baseline policy. 

\subsection{Storing Related Vectors Together}
If vectors that are accessed at short intervals from each other are also stored physically in the same blocks,
\name would be able to read fewer NVM blocks while fetching the same amount of data.
We explore two directions for physically placing the embedding vectors.

\emph{Semantic partitioning} assumes that two
vectors that are close in the Euclidian space may also share a semantic similarity, and thus should be stored together.
We use an unsupervised K-means clustering algorithm to evaluate this direction. 
\emph{Supervised partitioning} uses the history of past accesses to decide which vectors should be stored together.

In order to evaluate the benefits of physically storing related embedding vectors compared to the baseline policy, we start by
experimenting with an unlimited cache (i.e. a DRAM cache with no evictions), in which all blocks that are read are
cached. We calculate the effective bandwidth increase by comparing the number of block reads from the NVM using
a production workload of 1 billion requests.

\subsubsection{Semantic Partitioning with K-Means}

Our first hypothesis for deciding where to place vectors is that vectors that are close in the Euclidian space
are also accessed at close temporal intervals.
The intuition is that if vectors are close to each other, they might represent similar
types of content (e.g., similar pages), and would be accessed at similar times.

We can formalize the problem by expressing the embedding table $E$ as a $m \times n$ matrix
\begin{equation}
E = [\textbf{v}_{1},...,\textbf{v}_n]
\label{eq:emb}
\end{equation}
Suppose that $n$ embedding vectors are semantically meaningful, in other words, two vectors $\textbf{v}_i$ and $\textbf{v}_j$ that are close to each other in Euclidian distance $||\textbf{v}_i - \textbf{v}_j||_2$ sense are also more likely to be accessed at close temporal intervals. Then, we would like to find a column reordering $\textbf{p}$, such that
\begin{equation}
\min_p \sum_{i=0}^{n} || \textbf{v}_{p(i)} - \textbf{v}_{p(i+1)} ||_2
\end{equation}
We can approximate the solution to this problem by using K-means to cluster the vectors based on Euclidian distance,
and sort them so that vectors in the same cluster are ordered next to each other in memory.

\begin{figure}[t]
\begin{center}
  \includegraphics[width=0.9\columnwidth]{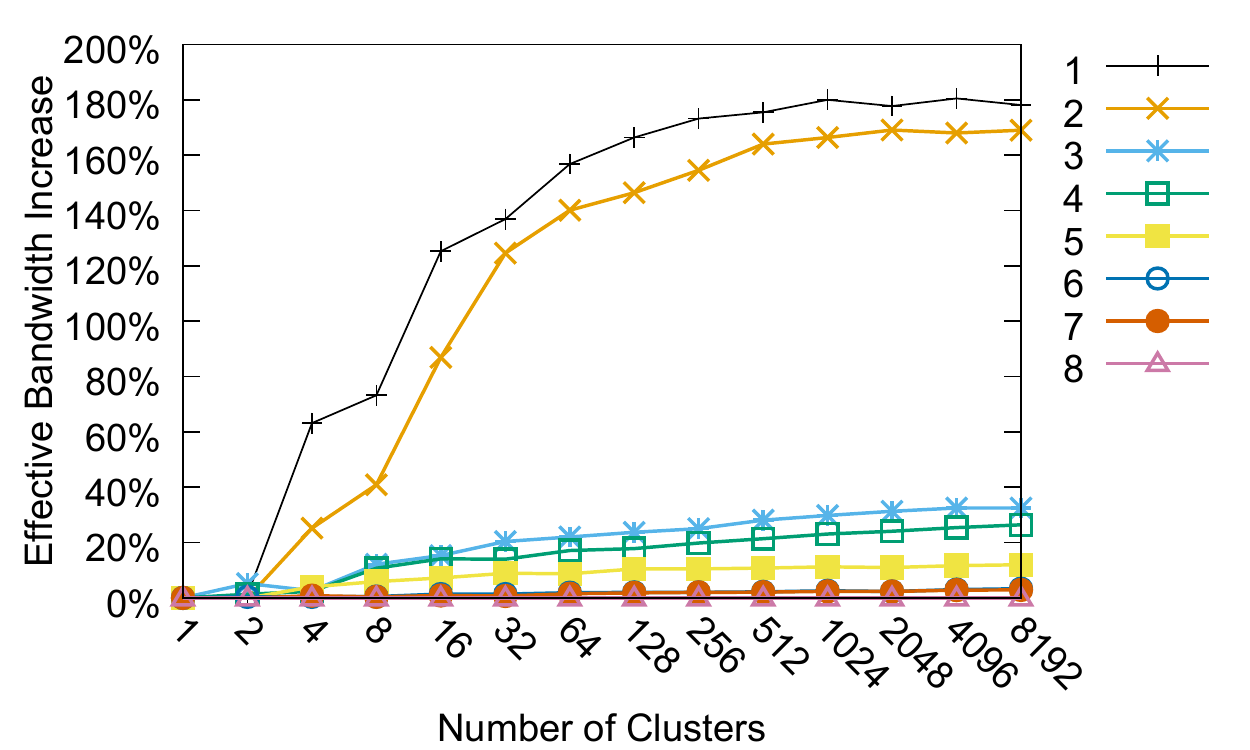}
  \vskip -0.1in
  \caption{Effective bandwidth increase when ordering embedding vectors according to their K-means clusters. The lines
represent different embedding tables.}
  \label{fig:kmeans_1level}
  \vskip -0.2in
  \end{center}
\end{figure}

\begin{figure*}[t]
 \begin{center}
  \subfigure[]{\label{fig:kmeans_1level_runtime}\includegraphics[width=.67\columnwidth]{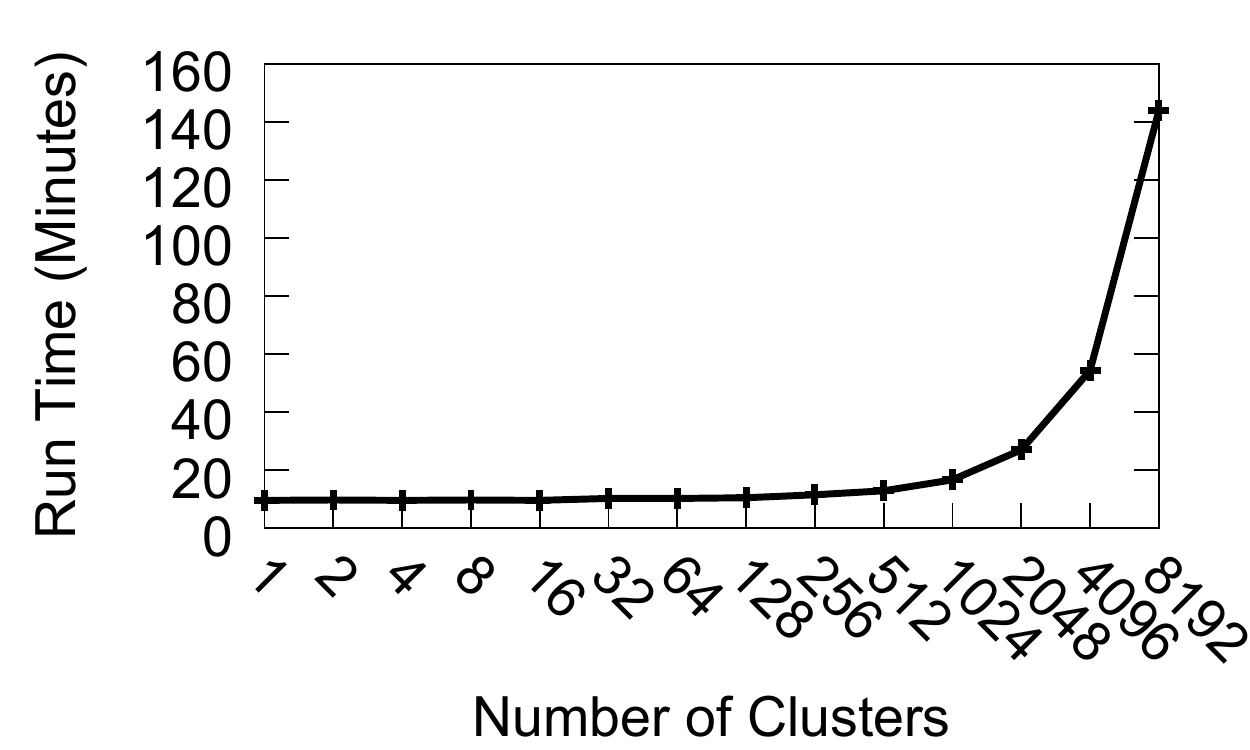}}
  \subfigure[]{\label{fig:kmeans_2levels_runtime}\includegraphics[width=.67\columnwidth]{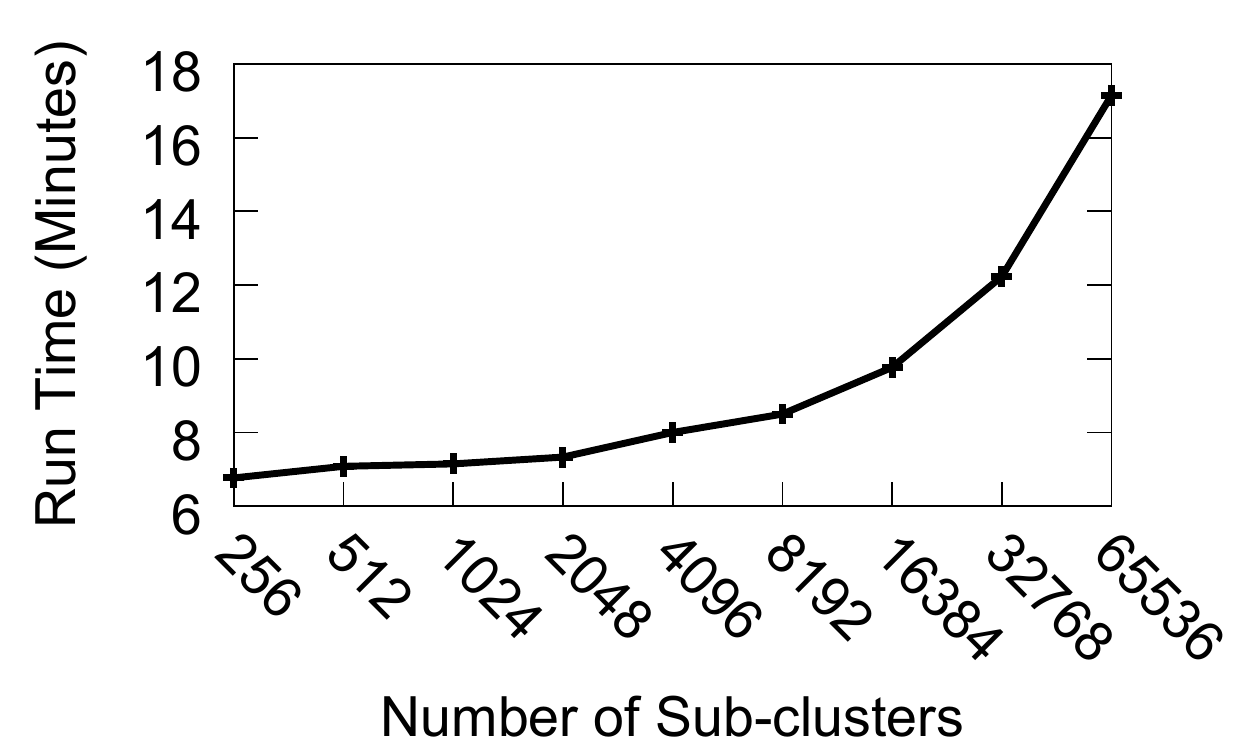}}
  \subfigure[]{\label{fig:shp_runtime}\includegraphics[width=.67\columnwidth]{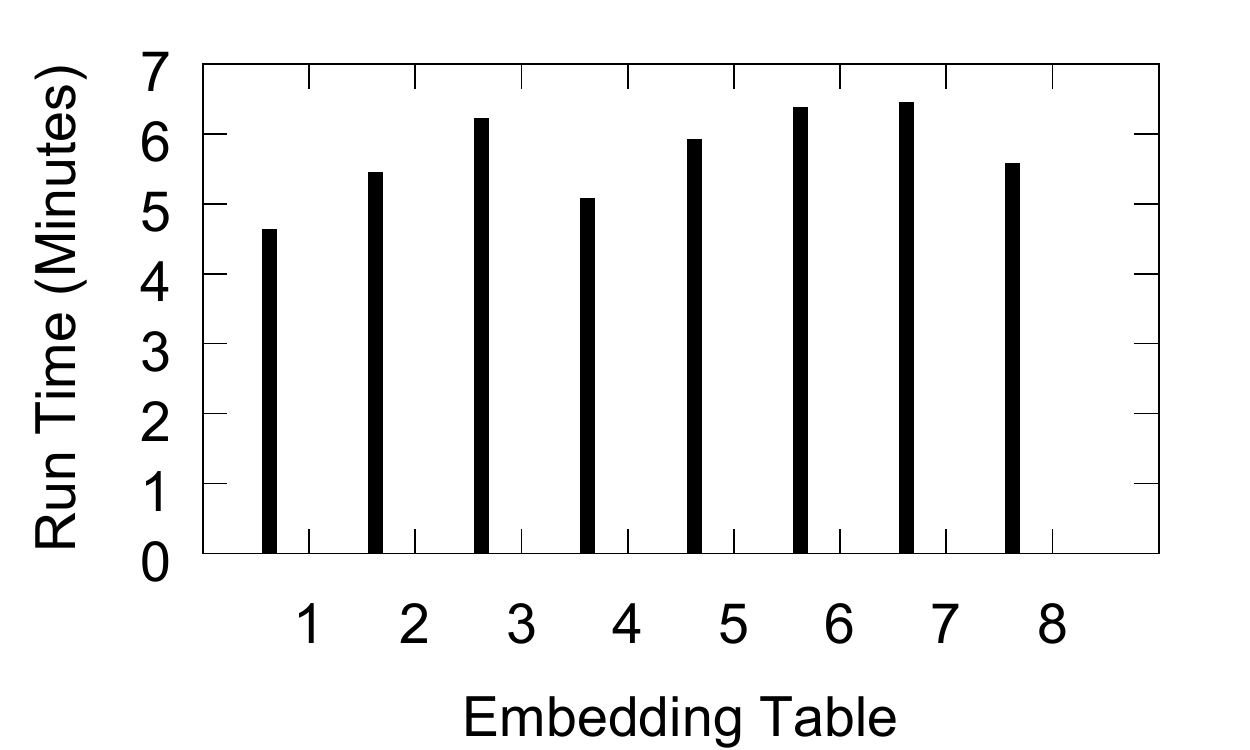}}
  \vskip -0.1in
  \caption{(a) The runtime of the K-means algorithm on embedding table 4, using the Faiss library  
      with 20 iterations and 24 threads. 
    (b) The runtime of two-stage K-means algorithm on embedding table 4, using the Faiss library  
      with 20 iterations and 24 threads. 
    (c) The runtime of the SHP algorithm  
            with 16 iterations and 24 threads per embedding table.}
 \vskip -0.3in
 \end{center}
\end{figure*}

Figure~\ref{fig:kmeans_1level} shows the effective bandwidth increase for different number of clusters.
The results show that for certain tables (e.g., tables 1 and 2), the effective bandwidth is increased significantly, while for others
it is not. Note that for example, table 8, which does not experience a large effective bandwidth increase, suffers from a high compulsory
miss rate (see Table~\ref{characterization-table}).
Ideally, we would use K-means with a large number of clusters.
Figure~\ref{fig:kmeans_1level_runtime} shows that the runtime of K-means increases exponentially as a function of the number of clusters.
Therefore, K-means does not scale to a large number of clusters (e.g., 625,000).

\begin{figure}[t]
\begin{center}
  \includegraphics[width=0.9\columnwidth]{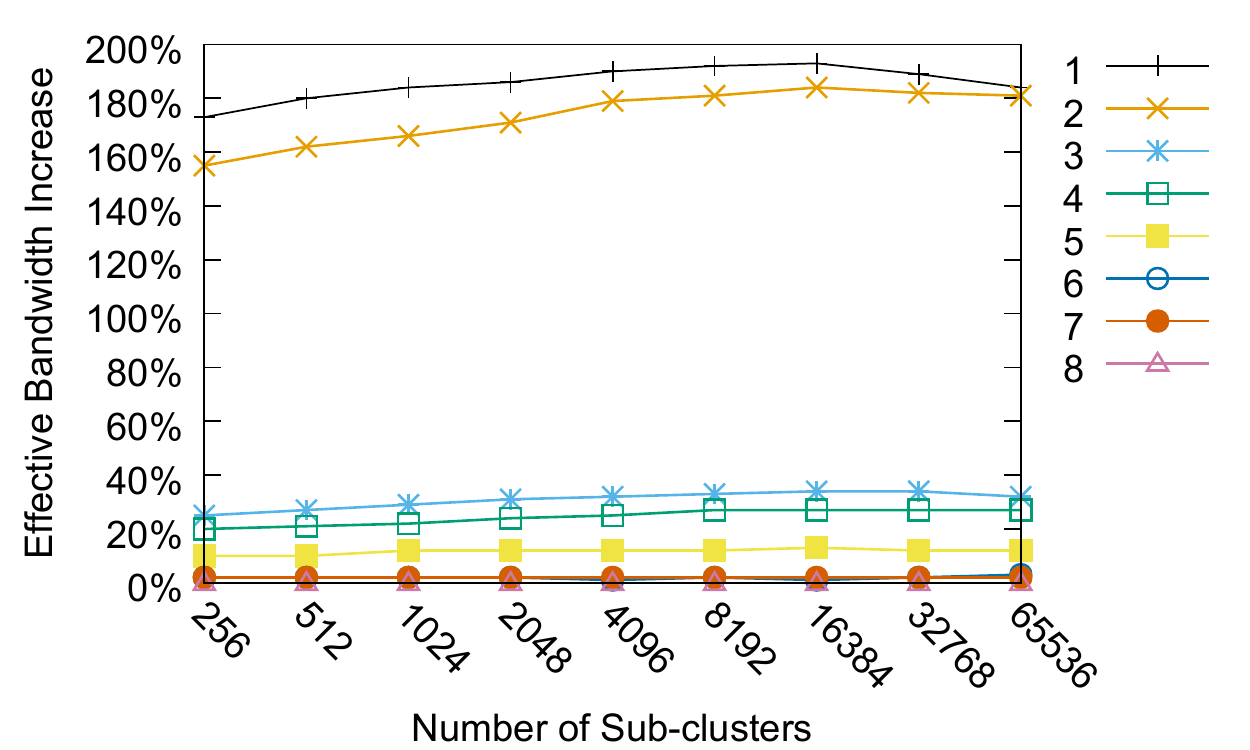}
  \vskip -0.1in
  \caption{Effective bandwidth increase when ordering the embedding vectors using recursive K-means. The lines represent different embedding tables.}
  \label{fig:kmeans_2levels}
  \vskip -0.3in
  \end{center}
\end{figure}

In order to reduce the runtime, we also experiment running an
appromixation of K-means by running the algorithm recursively. We first run K-means to cluster the embeddings into 256 clusters,
then recursively run it again on each of the clusters, creating ``sub-clusters''. 
Figure~\ref{fig:kmeans_2levels} depicts the effective bandwidth increase for different number of sub-clusters,
and Figure~\ref{fig:kmeans_2levels_runtime} shows the corresponding runtime as measured on table 4.
The results show that using recursive K-means does not reduce the effective bandwidth, and there is no benefit
increasing the number of clusters beyond 8,192.

\subsubsection{Supervised Partitioning with SHP}

In general, a small Euclidean distance of vectors in embedding tables does not always guarantee they are going to be
accessed together in time. Another problem of relying on the Euclidean distance, especially in the context of \company, is that
if vectors are frequently updated due to re-training (e.g., every hour), their Euclidean distance can change, and therefore
K-Means would need to be re-run at every update.
Therefore, we also experiment with an approach that does not rely on Euclidean distances,
but rather on the past access patterns of each vector. This approach would not require re-computing the classifier each time the vectors
get re-trained, since the identity of the vectors remains the same, even if their values get updated.

To do so, we are inspired by techniques introduced in the context of partitioning hypergraphs for optimizing data placement
for databases~\cite{zoltan,SHP}.
Suppose that we have a representative sequence of accesses to the embedding table,
which are represented as sparse IDs. Let these accesses be organized into lookup queries
$Q_j$, each corresponding to a particular user. Then, we would like to find a column
reordering $\textbf{p}$ such that columns accessed together by the same user are stored in the same block.

We can find a solution to this problem by mapping it to a hypergraph.
Let $H = (\mathcal{D},\mathcal{E})$ be a hypergraph with a set of vertices
$\mathcal{D}$ corresponding to sequence of accesses and a set of hyperedges
$\mathcal{E}$ corresponding to lookup queries $\mathcal{Q}_j$. Also, let
$p$ be a partitiong of the vertices $\mathcal{D} = \bigcup \mathcal{D}_i$ into disjoint blocks
$\mathcal{D}_i$ for $i=1,...,k$. Then, notice that the spatial locality of accesses can be
expressed as minimizing the \emph{average fanout}:
\begin{equation}
\min_p \frac{1}{n} \sum_{j=1}^{n} \left( \sum_{i=1}^{k} \texttt{intersect} (Q_j,\mathcal{D}_i) \right) 
\end{equation}
where \emph{fanout} in parenthesis is the number of blocks that need to be read to satisfy the query, with  
\begin{equation}
\texttt{intersect}(Q_j,\mathcal{D}_i) 
=
\left\{ 
\begin{array}{ll}
1 & \texttt{if } Q_j \bigcap \mathcal{D}_i \neq \emptyset \\
0 & \texttt{otherwise}
\end{array}
\right.
\end{equation}

Notice that the average fanout measures the number of blocks accessed by each query, rather than the general proximity of accesses.
Therefore, we temporally approximate vectors that are accessed in the \emph{same query}.
We can start with two blocks and apply the algorithm recursively on them~\cite{SHP}.

We configure the SHP algorithm blocks to contain 32 embeddings.
We first run the algorithm on a set of up to 5 billion requests, then measure the bandwidth reduction 
on a separate trace of 1 billion requests.

\begin{figure}[t]
\begin{center}
  \includegraphics[width=0.9\columnwidth]{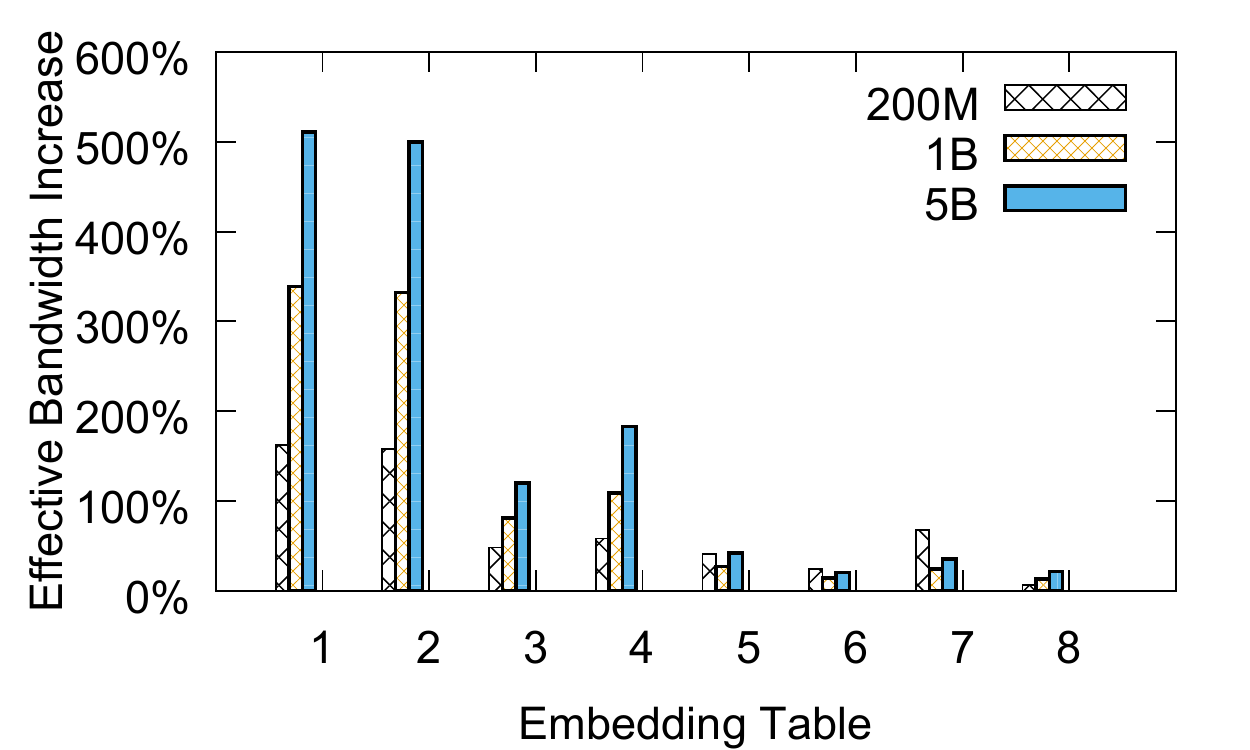}
  \vskip -0.1in
  \caption{Effective bandwidth increase when ordering the embedding vectors using SHP with an unlimited DRAM cache, as
	a function of the number of requests used to train SHP.}
  \label{fig:shp}
  \vskip -0.5in
  \end{center}
\end{figure}

Figure~\ref{fig:shp} depicts the effective bandwidth increase
per table, when running the SHP algorithm with different number of requests.
Overall, SHP exeeds the bandwidth savings achieved with K-means for all tables.
Utilizing larger datasets for running the algorithm improves its accuracy and the corresponding
effective bandwidth (we did not see a significant bandwidth improvement for using datasets larger than
5 billion requests). Figure~\ref{fig:shp_runtime} shows the SHP runtime per table, running with 16 iterations and 24 threads. 

\subsection{Caching the Embedding Tables}
\label{sec:caching}

So far, we assumed that \name uses an infinite cache. However, as we noted above,
the amount of DRAM we can dedicate to each table is limited. Therefore, \name
also needs to implement an eviction policy to decide which vectors to cache in DRAM. 
We experiment with using an eviction policy of Least Recently Used (LRU).

\begin{figure}[t]
\begin{center}
  \includegraphics[width=0.9\columnwidth]{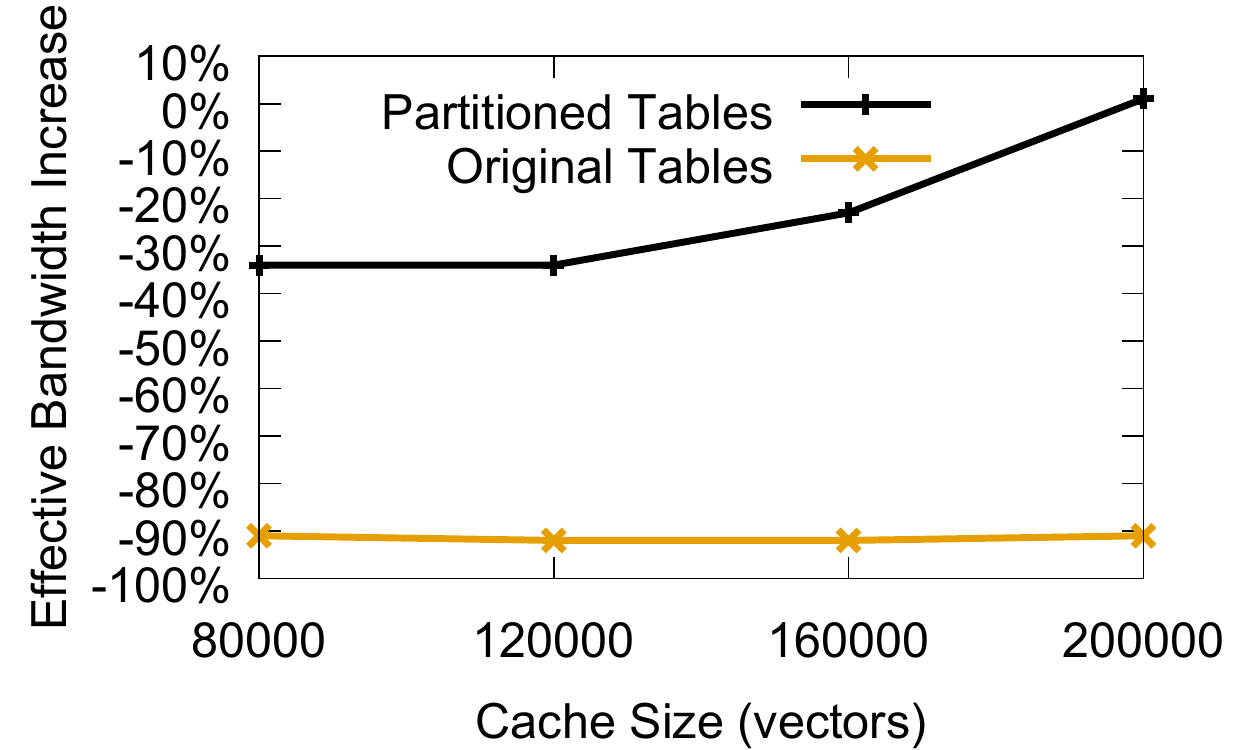}
  \vskip -0.1in
  \caption{Effective bandwidth increase when ordering the embedding vectors using SHP with a limited DRAM cache, with a policy
  of treating prefetched vectors the same as vectors that are read by the application. The figure also depicts
  the effective bandwidth of the unsorted original tables.}
  \label{fig:block_prefetching}
  \vskip -0.3in
  \end{center}
\end{figure}

The first question when deciding which vectors to cache in DRAM, is how to treat the vectors that were
pre-fetched when \name reads the whole block.
Figure~\ref{fig:block_prefetching} depicts the effective bandwidth increase when all of the vectors in a block are cached and treated the same as the actual
requested vector for the original tables and for the partitioned tables. As the figure shows, simply allocating all 32 vectors in a block to the cache will trigger 32 evictions of potentially 
more useful vectors that are ranked higher in the eviction queue, reducing the cache hit rate and reducing the effective bandwidth significantly.
 
\subsubsection{Caching the Prefetched Vectors}
\begin{figure*}[t!]
 \begin{center}
  \subfigure[]{\label{fig:insertion_position_14}\includegraphics[width=.67\columnwidth]{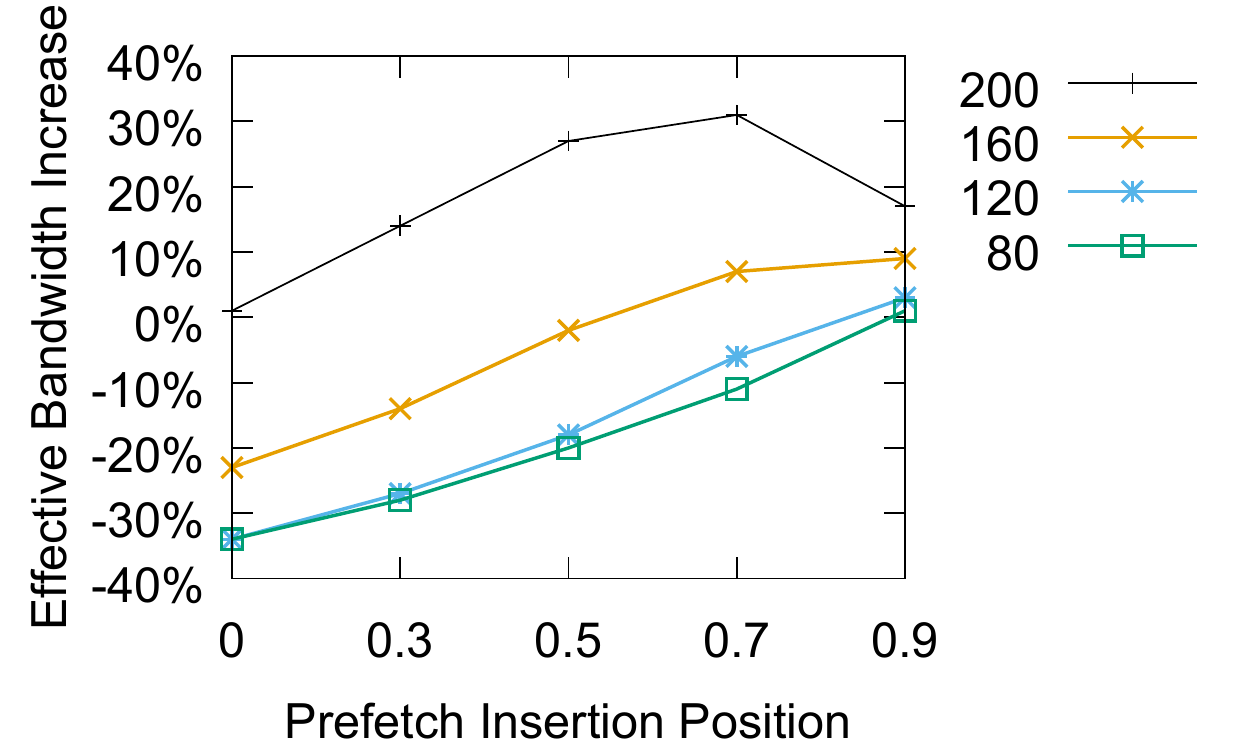}}
  \subfigure[]{\label{fig:shadow_queue_14}\includegraphics[width=.67\columnwidth]{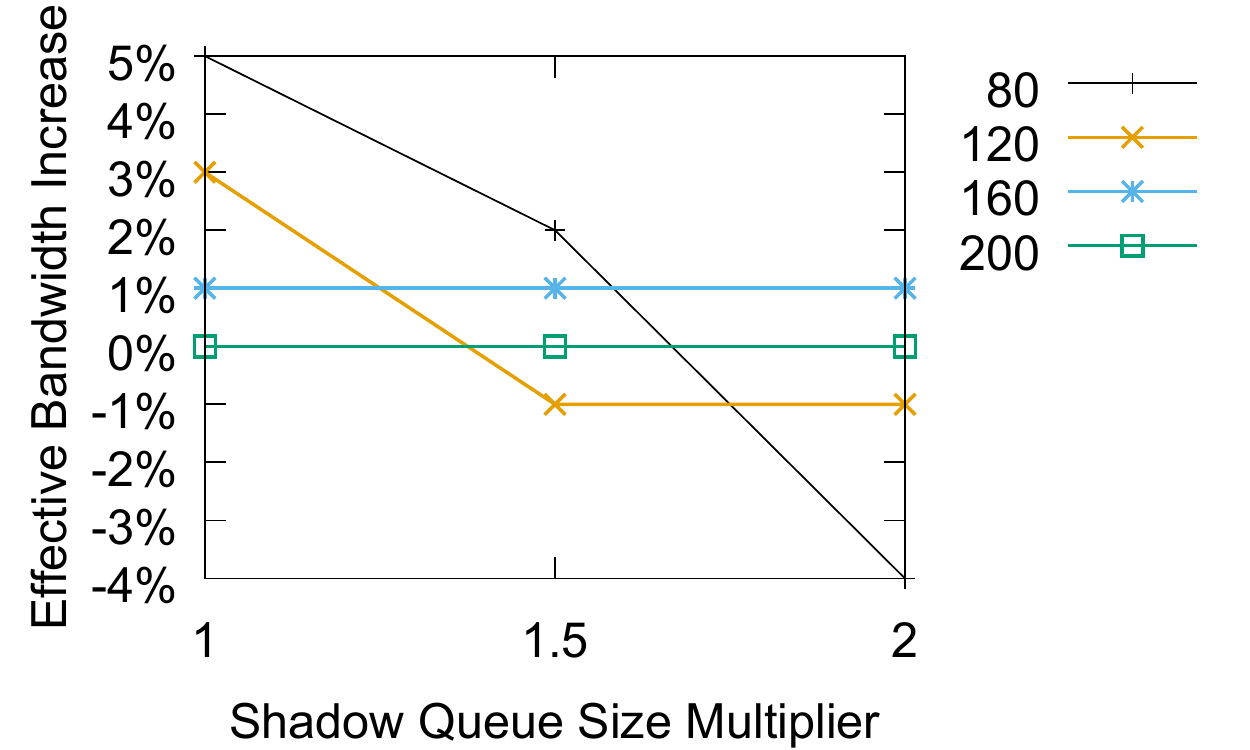}}
  \subfigure[]{\label{fig:insertion_shadow_combined_14}\includegraphics[width=.67\columnwidth]{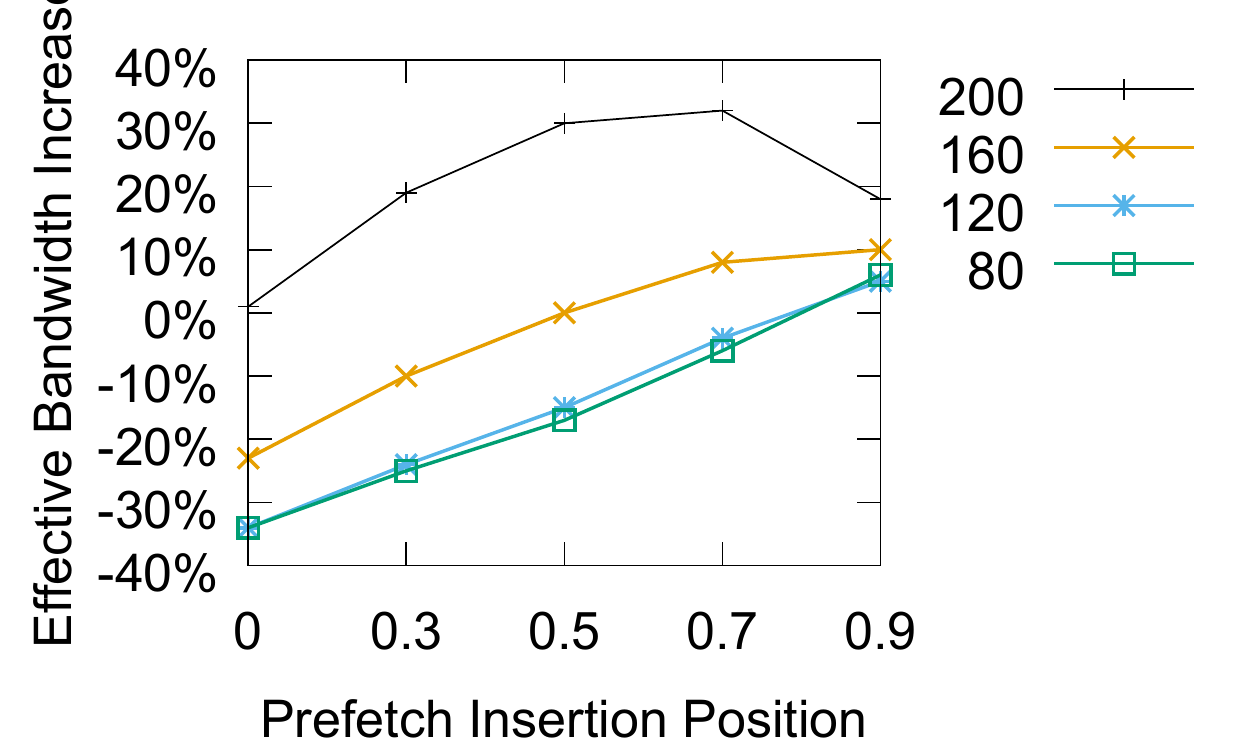}}
  \vskip -0.1in
  \caption{(a) Effective bandwidth increase when inserting prefetches to a different position in the queue, compared with no prefetching.
		(b) Effective bandwidth increase when filtering prefetches based on the shadow queue, compared with no prefetching.
		(c) Effective bandwidth increase when combining both methods, compared with no prefetching.
		The lines in all figures represent different cache sizes (vectors$\times$$10^{3}$).}
    \vskip -0.3in
 \end{center}
\end{figure*}

Even though our ordering algorithms improve the spatial locality of blocks, some prefetched vectors are not
accessed at all. This led us to experiment 
with inserting them in different positions in the eviction queue. Inserting prefetched vectors at a lower 
position in the queue prevents them from triggering the eviction of hot vectors, but also may shorten 
their lifetime in the cache and make them less likely to get accessed before they are evicted, thus
decreasing the effective bandwidth. 
Overall, we noticed that while improving the hit rate and bandwidth (compared with inserting prefetches at the top of the queue),
this method did not significantly affect the performance 
for lower cache sizes, and still provided low (and sometimes negative) bandwidth benefits. The main reason for this
is that all prefetched vectors are still allocated to the cache, without filtering the less useful vectors.
Figure~\ref{fig:insertion_position_14} presents the bandwidth reduction in embedding table 2 when inserting
prefetched vectors to different positions in the queue, over a baseline with no prefetches. The X axis  
represents the prefetch insertion position relative to the top of the eviction queue (e.g. 0.5 and 0  mean the 
middle and top of the queue, respectively). 

Instead of allocating prefetched vectors to a lower point in the queue, we can use an admission control algorithm
that decides whether prefetched vectors enter the queue at all.
As a first step, we use a separate LRU list, which we term a \emph{shadow cache}, which stores only the index of the vectors, without storing their content.
We allocate only vectors that were explicitly read, thus simulating another 
cache (that has no prefetched vectors) without actually caching the value of the embedding vectors. 
When a block is read from the NVM, its vectors are prefetched only if they already appear in the shadow cache 
(note that the vector read by the application is always cached). Figure~\ref{fig:shadow_queue_14} demonstrates the effective 
bandwidth increase as a function of the size of the shadow cache with table 2.  
The shadow cache size is calculated using a multiplier over the real cache size. 
For example, a mutiplier of 1.5 for a cache of 80,000 vectors means that the shadow cache size 
 is 120,000 vectors.  

As shown in Figure~\ref{fig:shadow_queue_14}, the shadow cache produces a very small effective bandwidth increase when used as an admission policy. 
The existence of a vector in the shadow cache does not correlate with its usefulness as a prefetched vector. 
We try to combine both methods, by using the shadow cache to decide where to allocate the prefetched vector in queue. If the prefetched vector hits in the
simulated cache, it is allocated to the top of the queue. Otherwise, it is allocated to the separate insertion position. 
Figure~\ref{fig:insertion_shadow_combined_14} shows the effective bandwidth increase in embedding table 2 when using this method, over a baseline with no prefetches.

\subsubsection{Dealing with Rarely Accessed Vectors}
We try to improve bandwidth savings further by leveraging the insight that during SHP run, some vectors are rarely accessed,
as demonstrated in Figure~\ref{fig:accesses}.
SHP has very limited information about such vectors and on how to sort them. However, since SHP performs balanced partitioning, all the 
vectors are paritioned to equally-sized blocks, and blocks may contain vectors that were rarely (or never) accessed during the SHP run.
In such cases SHP will simply assign them to arbitrary locations in blocks that have free space.

Driven by this insight, \name collects statistics on the number of times each vector was accessed during the SHP run (i.e, how many queries contained 
each vector). When reading a 4~KB block from NVM, vectors will be prefetched only if they were accessed $>$ \emph{t} times during the SHP run.
Figure~\ref{fig:threshold} presents the effective bandwidth increase with different threshold values \emph{t} for table 2, compared to a baseline of no prefeches.
This policy significantly improves effective bandwidth.
The number of vector accesses during an SHP run correlates with their utility as prefetched vectors, since SHP has more 
confidence in assigning a useful location for vectors that appeared in many queries. For smaller cache sizes, the price of cache evictions is higher,
hence \name should utilize higher thresholds to filter out the more speculative prefetches.
For higher cache sizes, \name should use lower thresholds to more aggressively prefetch.

\subsubsection{Configuring the Cache Parameters with Simulations}
\begin{figure}[t]
 \begin{center}
  \includegraphics[width=0.8\columnwidth]{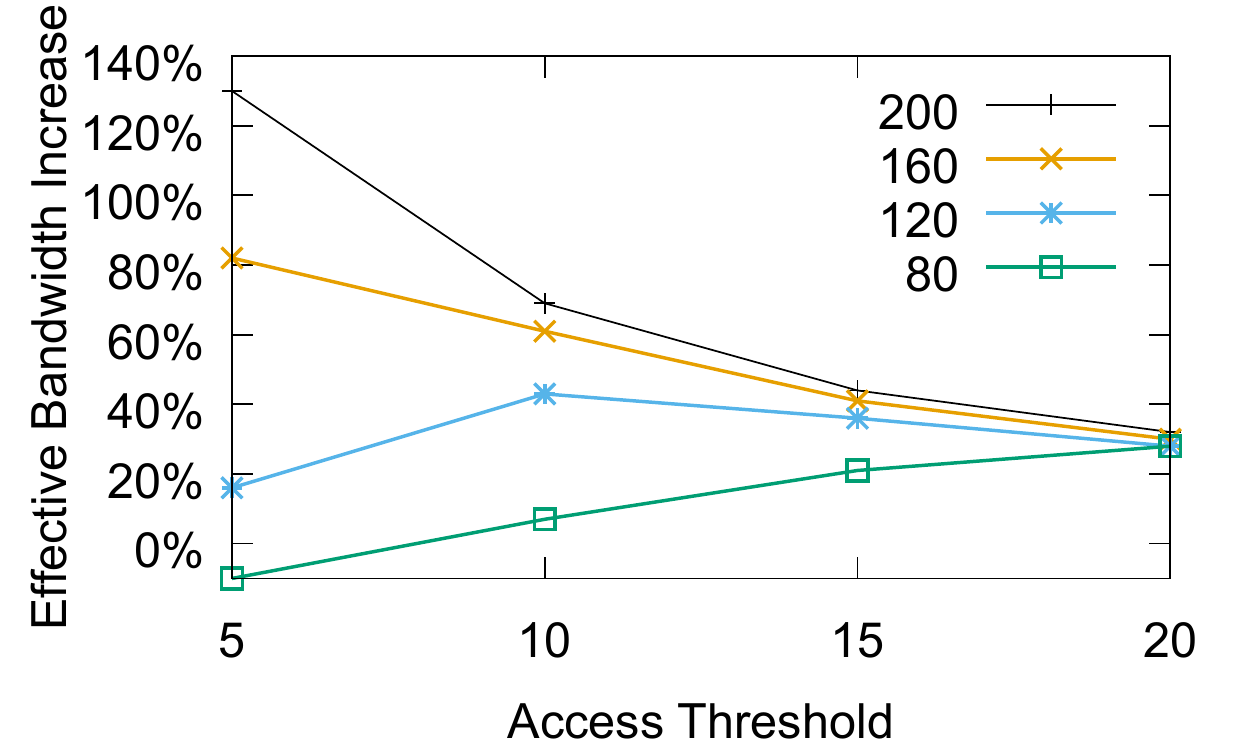}
  \vskip -0.1in
  \caption{Effective bandwidth increase when filtering prefetched vectors based on the number of accesses during SHP run. The lines represent different cache sizes (vectors$\times$$10^{3}$).}
  \label{fig:threshold}
  \vskip -0.3in
 \end{center}
\end{figure}

As Figure~\ref{fig:threshold} shows, the optimal threshold varies across different  
cache sizes. Picking an a-priori one-size-fits-all
threshold for all the tables would lead to a low effective bandwidth. Ideally, \name should automatically pick the right threshold for each table
and cache size and automatically
tune it for each table and cache size.

To do so, we borrow an idea used from key-value caches, called ``miniature caches''~\cite{mini-caches}.
The idea behind miniature caches is to simulate the hit rate curve of multiple different cache configuration, or in our case,
simulate the cache with different thresholds for prefetched vectors, and pick the one that provides the highest
hit rate for a given cache size. The main problem with this approach is how to simulate multiple caches in real-time
without incurring a high performance and memory overhead.

Miniature caches uses the insight that hit rate curves can be estimated efficiently without having to use the entire
access workload, but rather by randomly sampling requests from the workload and computing the hit rate curve for the sampled requests.
For example, if the total cache is of size $S$,
and we sample the request stream at a rate of $\frac{1}{N}$, the miniature cache only needs to track
$\frac{S}{N}$ vectors. In addition, the miniature cache does not have to store the value of the objects,
only their IDs.

In our case, we find that in order to accurately simulate a cache, we can down-sample its requests
by a factor of 1000.
Table~\ref{mini-sim-table} compares the ideal threshold when running embedding table 2 with different cache sizes,
to the thresholds chosen by the miniature cache simulations with different sampling rates. The results show
that there is not a big difference in the effective bandwidth between the ideal thresholds and the ones chosen by the simulations.
It also shows that for larger caches, \name can use a more relaxed threshold, while for smaller caches with less
DRAM, \name benefits from using a more aggressive admission control policy for prefetched vectors.

In fact, the hit rate curves produced by the miniature caches not only provide the ideal threshold for prefetched vectors, but also
allow the datacenter operator to optimize the amount of DRAM across the different tables to maximize performance.
There are various techniques for maximizing total hit rate across multiple hit rate curves, including when the curves are convex~\cite{dynacache,memshare},
and even when they are not convex~\cite{Talus,cliffhanger,mini-caches}. 
In our case, we found that the hit rate curves of all the tables
are convex and do not change substantially across runs. Therefore, we ran \name on a trace with 5 billion requests and
statically assigned the amount of DRAM to assign to each table with the goal of optimizing the total hit rate~\cite{dynacache}.

\begin{table*}[t]
\caption{Measuring the effectiveness of using miniature caches with different sampling rates with embedding table 2. On the left, the results show the ideal
admission control threshold for the full cache, for different cache sizes.
The results to the right show the chosen threshold when running miniature caches, using different sampling ratios.
Even at 0.1\% sampling, miniature caches provides a relatively similar bandwidth gain compared to the ideal threshold.}
\label{mini-sim-table}
\vskip -0.1in
\begin{center}
\begin{small}
\begin{sc}
\setlength{\tabcolsep}{0.3em}
\begin{tabular}{l||cc|cc|cc|cc}
\toprule
    Size   & \multicolumn{2}{l|}{Full Cache} & \multicolumn{2}{l|}{10\% Sampling} & \multicolumn{2}{l|}{1\% Sampling} & \multicolumn{2}{l}{0.1\% Sampling} \\
\midrule  
      & Threshold     & BW gain    & Threshold      & BW gain     & Threshold     & BW gain     & Threshold     & BW gain      \\ 

80,000  &     20      &   27.6\%    &    20         &    27.6\%  &      15      &   21.4\%   &     15       &   21.4\%     \\ 
120,000 &     10      &   43.0\%    &    15         &    36.3\%  &      10      &   43.0\%   &      15       &  36.3\%     \\ 
160,000 &      5       &  80.3\%    &     5          &    80.3\%  &       5      &   80.3\%   &      10       &  61.0\%     \\ 
200,000 &      5       &  129.9\%    &     5          &    129.9\%  &       5      &  129.9\%   &       5        &  129.9\%     \\ 
\bottomrule
\end{tabular}
\end{sc}
\end{small}
\end{center}
\vskip -0.3in
\end{table*}

\section{End-to-End Evaluation}

\begin{figure}[t]
 \begin{center}
  \includegraphics[width=0.9\columnwidth]{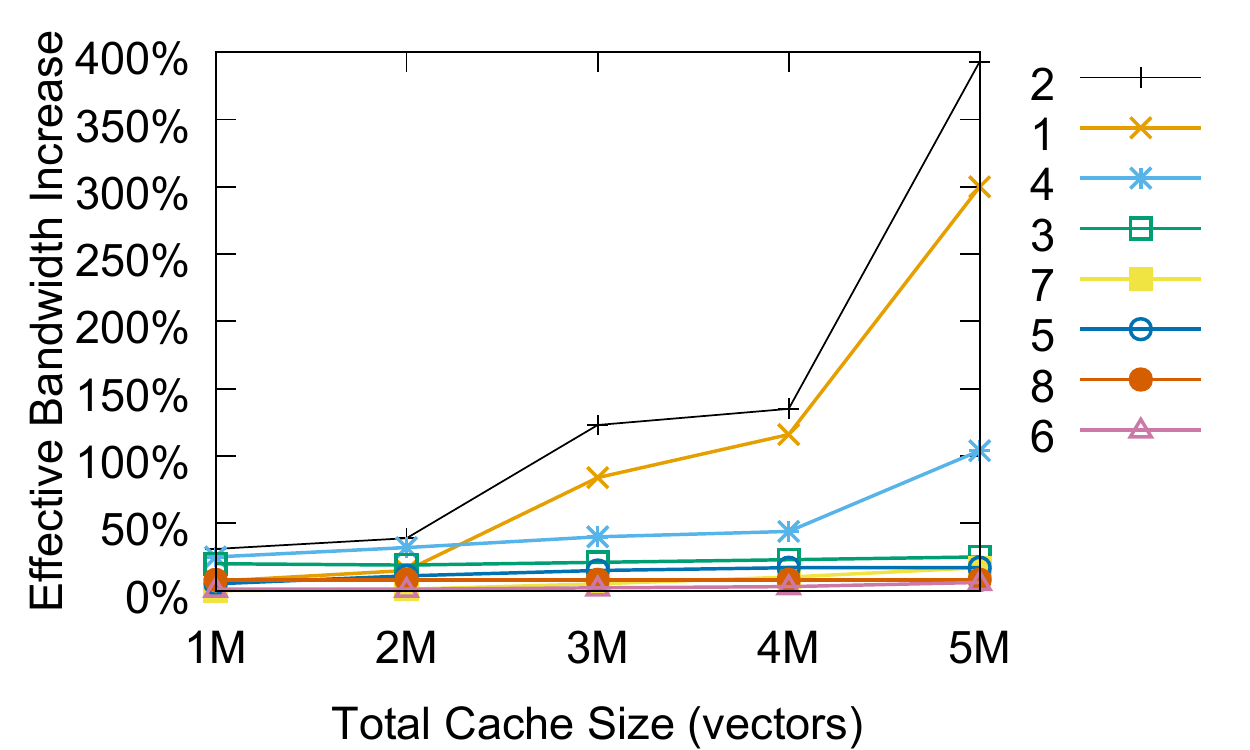}
  \vskip -0.1in
  \caption{Effective bandwidth increase as a function of total cache size. The lines represent different embedding tables.}
  \vskip -0.2in
  \label{fig:eval_cache_sizes}
 \end{center}
\end{figure}

\begin{figure}[t]
 \begin{center}
  \includegraphics[width=0.9\columnwidth]{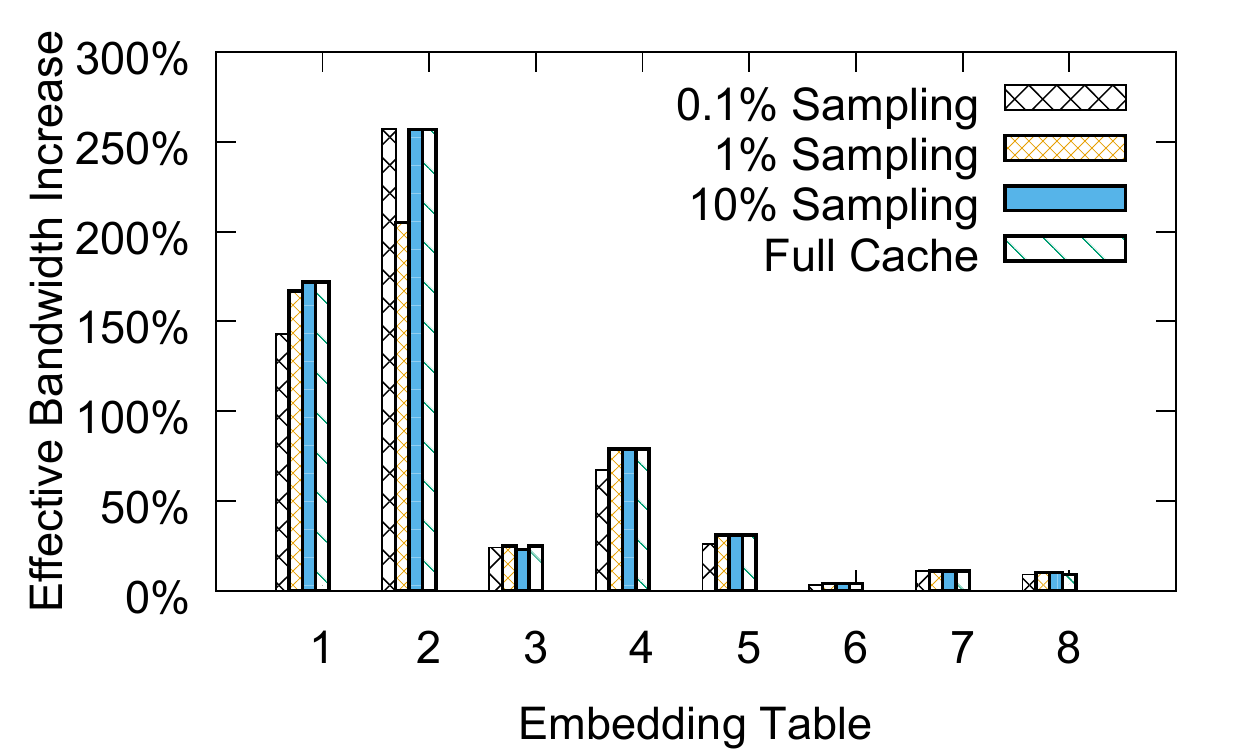}
  \vskip -0.1in
  \caption{Effective bandwidth increase as a function of the sampling rate of the miniature caches, trained on 5 billion requests. The full cache policy represents an oracle
  policy that selects the ideal threshold for each table.}
  \vskip -0.1in
  \label{fig:eval_sampling}
 \end{center}
\end{figure}

\begin{figure}[t]
 \begin{center}
  \includegraphics[width=0.9\columnwidth]{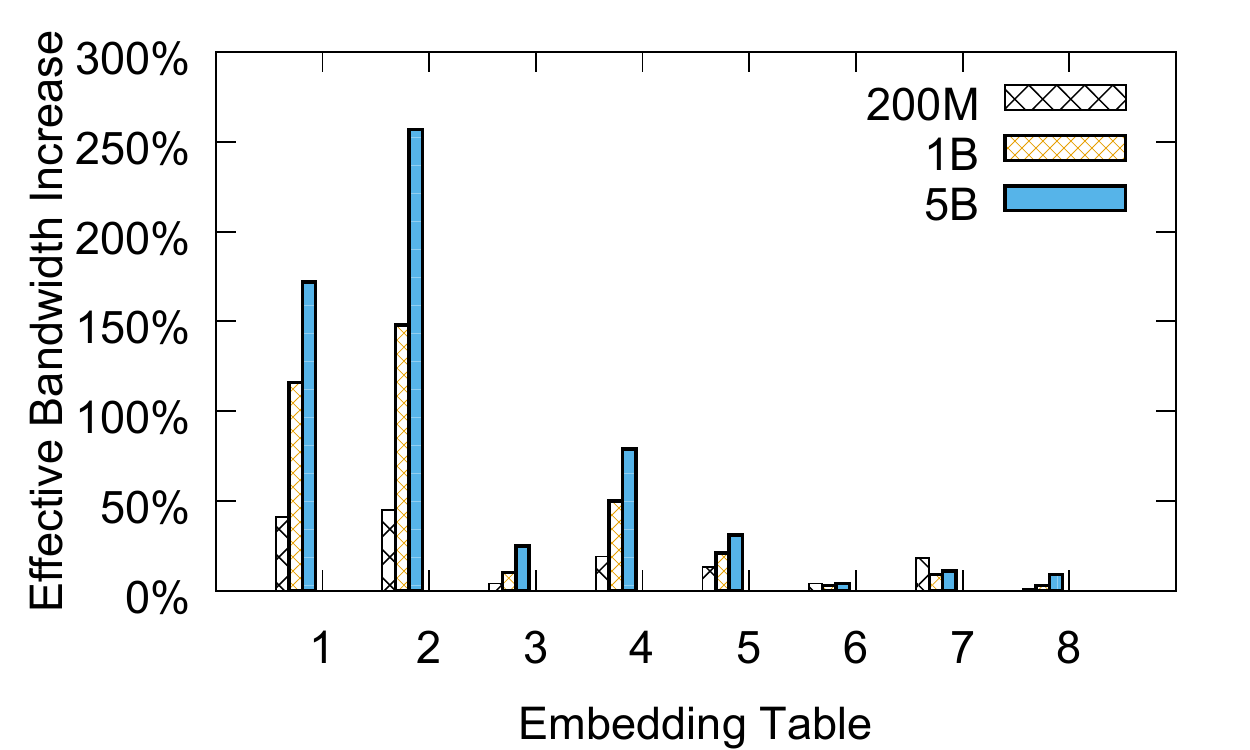}
  \vskip -0.1in
  \caption{Effective bandwidth increase as a function of the number of requests used to train SHP, evaluated against 1 billion requests.}
  \vskip -0.2in
  \label{fig:eval_training}
 \end{center}
\end{figure}

\begin{figure}[t]
 \begin{center}
  \includegraphics[width=0.9\columnwidth]{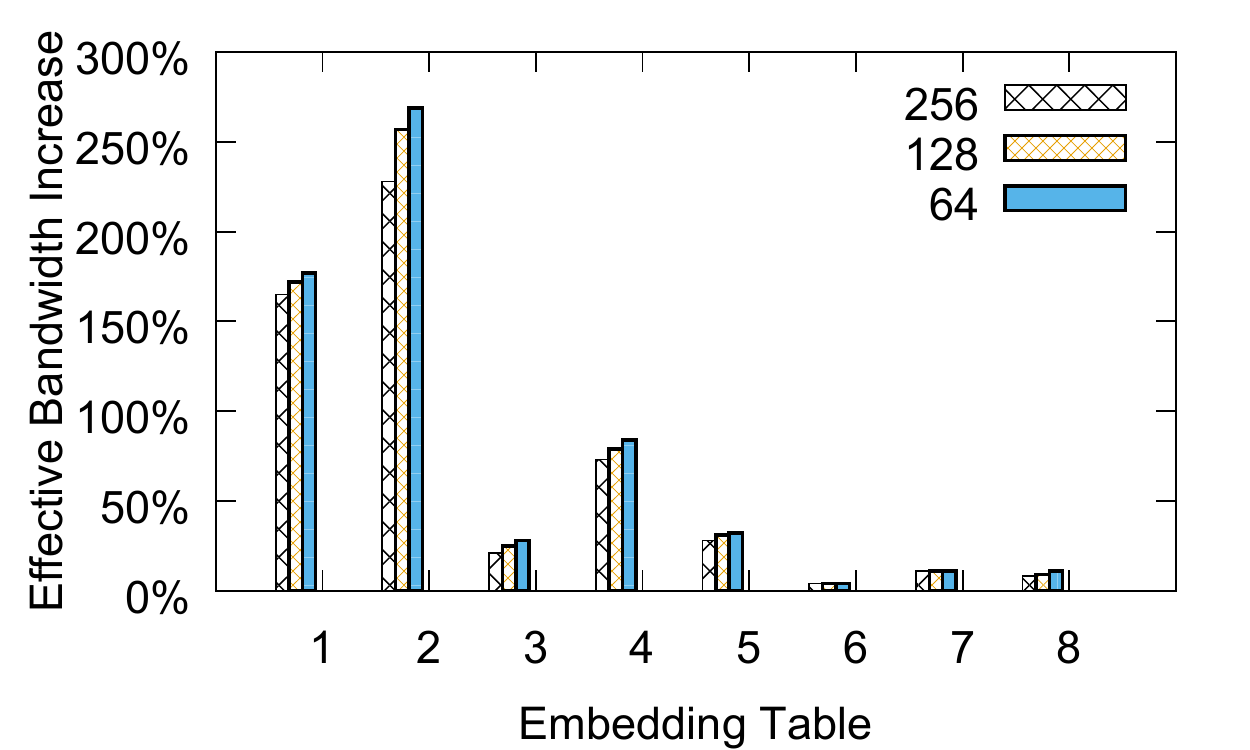}
  \vskip -0.1in
  \caption{Effective bandwidth increase as a function of the embedding vector size (bytes).}
  \vskip -0.2in
  \label{fig:eval_vector_sizes}
 \end{center}
\end{figure}

In this section we analyze the end-to-end effective bandwidth increase of \name under different scenarios:
(1) as a function of cache size, (2) as a function of the simulated cache size, (3) as a function of SHP's training data size,
and (4) as a function of the embedding vector size.
In all the experiments in this section, unless otherwise specified, we ran \name using SHP trained on 5 billion requests, with a total cache of 4 million vectors, simulated caches with a size of 0.1\% of the total cache, and a vector size of 128~B.

Figure~\ref{fig:eval_cache_sizes} compares the total effective bandwidth increase across 8 different tables as a function of cache size.
The graph shows that for certain tables, the effective bandwidth significantly increases as a function of the cache size,
up to almost 5$\times$ for table 2. For some tables however, the effective bandwidth remains relatively stable and low.
The reason for this is that the access patterns of some tables are simply more random, and harder to effectively partition and cache.
At the extreme, an application that accesses embedding vectors completely uniformly would not see any effective bandwidth increase.

We also analyze the impact of the size of the miniature caches in Figure~\ref{fig:eval_sampling}.
The figure shows that the effective bandwidth is almost the same with an oracle policy that selects the ideal prefetched vector threshold,
compared to a miniature cache simulation that is scaled down to a thousandth of the size of the cache.

Figure~\ref{fig:eval_training} varies the number of training samples. It shows that as we increase the training time, the effective bandwidth increases.
This is due to the fact that SHP's effectiveness in placing related vectors physically together improves as a function of the amount of training data.

While our model currently uses a vector size of 128 bytes, we also compare the effective bandwidth increase for
different vector sizes in figure~\ref{fig:eval_vector_sizes}, using a total cache of 4 million vectors (i.e. the cache size changes proportionally with the vector size). When vector sizes are smaller, each NVM block accomodates more
vectors, enabling \name to achieve higher effective bandwidth increase. 

\section{Related Work}

\name uses techniques inspired by prior research in using NVM as a substitute for DRAM, 
partitioning and caching. 

NVM has been proposed as a low cost substitute for DRAM in other contexts, including databases and file systems.
MyNVM~\cite{MyNVM} is a SQL database based on MyRocks, which uses block-level NVM as a second level cache
for flash, and a lower cost substitute for DRAM. Similar to \name, one of the main challenges MyNVM
deals with is NVM's limited bandwidth compared to DRAM. However, unlike \name, MyNVM stores
objects in relati-vely large files (e.g., 4-6~KB). The novel challenge addressed by \name is how to physically place
and cache embeddings vectors, which are much smaller than NVM blocks.

Other databases simulate NVM in its byte-addressable form, such as:
CDDS~\cite{Venkataraman:2011}, Echo~\cite{Bailey:2013}, FPTree~\cite{Oukid:2016}, and HiKV~\cite{HiKV}.
There has also been several prior projects in using NVM in its byte-addressable form for
file systems, including: NOVA-Fortis~\cite{NOVA}, LAWN~\cite{LAWN}, and ByVFS~\cite{ByVFS}.
All of these systems use simulations to estimate how byte-addressable NVM will perform.
Unfortunately since byte-addressable
NVM is not commercially available, its real performance
characteristics are unknown.

\name's mechanism for ordering vectors in physical blocks uses SHP, a
hypergraph partitioning algorithm originally proposed for database query optimization~\cite{SH,SHP}.
The reason we chose SHP is due to its scalability and ease of implementation.
Another hypergraph partitioning algorithm used for database query partitioning is SWORD~\cite{SWORD}.
Zoltan~\cite{zoltan} and Parkway~\cite{parkway} are other distributed hypergraph partitioning algorithms.
However, both algorithms do not scale well for partitioning large workloads in our application (for more details see SHP~\cite{SHP}).

\name uses micro-simulations to test different cache admission thresholds. A similar approach
was used in recent work~\cite{mini-caches} to approximate miss-rate curves of different caching algorithms,
and select the optimal in real-time. Talus~\cite{Talus} and Cliffhanger~\cite{cliffhanger} demonstrate how miss-rate
curves can be estimated by simulating a small cache.
Other recent low cost miss-rate curve approximation techniques include
Counter Stacks~\cite{CounterStacks}, SHARDS~\cite{SHARDS}, and AET~\cite{AET}.

\section{Conclusions}

\name is a novel NVM-based storage system for storing deep learning models. It provides a lower cost
alternative to existing fully DRAM-based storage. \name reorders embedding vectors and stores
related ones physically together for efficient prefetching, and dynamically adjusts its caching
policy by simulating miniature caches for each embedding table. Similar techniques employed by \name
can be extended for using NVM to store other types of datasets, which require granular access to data.

\bibliography{bib}
\bibliographystyle{sysml2019}

\end{document}